\documentclass{article} % For LaTeX2e
\usepackage{iclr2027_conference,times}

\usepackage{amsmath,amsfonts,bm}

\def\eqref#1{equation~\ref{#1}}
\def\1{\bm{1}}

\DeclareMathAlphabet{\mathsfit}{\encodingdefault}{\sfdefault}{m}{sl}
\SetMathAlphabet{\mathsfit}{bold}{\encodingdefault}{\sfdefault}{bx}{n}

\usepackage{hyperref}
\usepackage{url}

\usepackage{tabularx}
\usepackage{array}
\usepackage{amsmath}
\usepackage{svg}
\usepackage{graphicx}
\usepackage{subfigure}
\usepackage{enumerate}
\usepackage{enumitem}
\usepackage{algorithm}
\usepackage{algpseudocode}
\usepackage{amssymb}
\usepackage{multirow}
\usepackage{longtable}
\usepackage{bm}
\usepackage{circuitikz}
\usepackage{wrapfig}
\usepackage{tabulary}
\usepackage{soul}
\sethlcolor{red}
\usepackage[utf8]{inputenc} % allow utf-8 input
\usepackage[T1]{fontenc}    % use 8-bit T1 fonts
\usepackage{hyperref}       % hyperlinks
\usepackage{url}            % simple URL typesetting
\usepackage{booktabs}       % professional-quality tables
\usepackage{amsfonts}       % blackboard math symbols
\usepackage{nicefrac}       % compact symbols for 1/2, etc.
\usepackage{microtype}      % microtypography
\usepackage{xcolor,colortbl}         % colors
\definecolor{lightergray}{HTML}{E9E9E9}
\definecolor{lighterpurple}{HTML}{E4DDF3}
\definecolor{darkgreen}{RGB}{50,100,0}
\definecolor{darkred}{RGB}{200, 0, 0}
\definecolor{lightblue}{RGB}{220,235,250}
\definecolor{red}{RGB}{184, 84, 80}
\definecolor{green}{RGB}{130, 179, 102}
\usepackage[table]{xcolor}
\definecolor{rowVtwoSixty}{HTML}{F7EDFD}
\definecolor{rowVtwoNinety}{HTML}{E1D5E7}
\usepackage[most,skins,theorems]{tcolorbox}
\tcbset{
  takeawaysbox/.style={
    colback=lighterpurple!50,
    colframe=black,
    width=\linewidth,
    arc=2.0mm,
    boxrule=0.6pt,
  }
}

\definecolor{casequestionbg}{HTML}{F2F7FC}
\definecolor{casequestiontitle}{HTML}{DCEBFA}
\definecolor{caseoutputtitle}{HTML}{EAE4F6}
\definecolor{casepurple}{HTML}{6F42C1}
\definecolor{casepurplebg}{HTML}{F3EEFC}
\definecolor{casegray}{HTML}{5F6673}
\definecolor{caseborder}{HTML}{C8CED8}
\definecolor{casegreen}{HTML}{147D64}
\definecolor{casegreenbg}{HTML}{ECF8F4}
\definecolor{anglebin1}{HTML}{FAF2E9}
\definecolor{anglebin2}{HTML}{F5EDE9}
\definecolor{anglebin3}{HTML}{F0E8E8}
\definecolor{anglebin4}{HTML}{EBE2E7}
\definecolor{anglebin5}{HTML}{E7DDE7}
\definecolor{anglebin6}{HTML}{E2D8E6}

\newcommand{\caseRawTag}[1]{%
  \textcolor{casegray}{\texttt{\detokenize{#1}}}%
}
\newcommand{\caseLatentSpan}[1]{%
  \tcbox[
    on line,
    enhanced,
    nobeforeafter,
    boxrule=0.7pt,
    arc=1.1mm,
    boxsep=0pt,
    left=1.0mm,
    right=1.0mm,
    top=0.45mm,
    bottom=0.45mm,
    colback=casepurplebg,
    colframe=casepurple
  ]{\textcolor{casepurple}{\bfseries\ttfamily\footnotesize\detokenize{#1}}}%
}
\newcommand{\caseExplicitEllipsis}{%
  \textcolor{casegray}{\itshape\ldots}%
}

\definecolor{nblue}{RGB}{41, 52, 190}
\hypersetup{
    colorlinks=true,   % 开启链接颜色
    linkcolor=purple,  % 内部链接颜色
    citecolor=nblue,   % 参考文献引用颜色
    filecolor=magenta, % 文件链接颜色
    urlcolor=nblue     % URL链接颜色
}
\def\github{\raisebox{-1.5pt}{\includegraphics[height=1.05em]{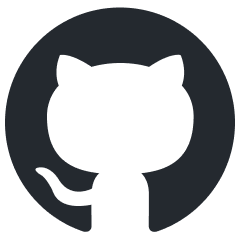}}}
\def\huggingface{\raisebox{-1.5pt}{\includegraphics[height=1.05em]{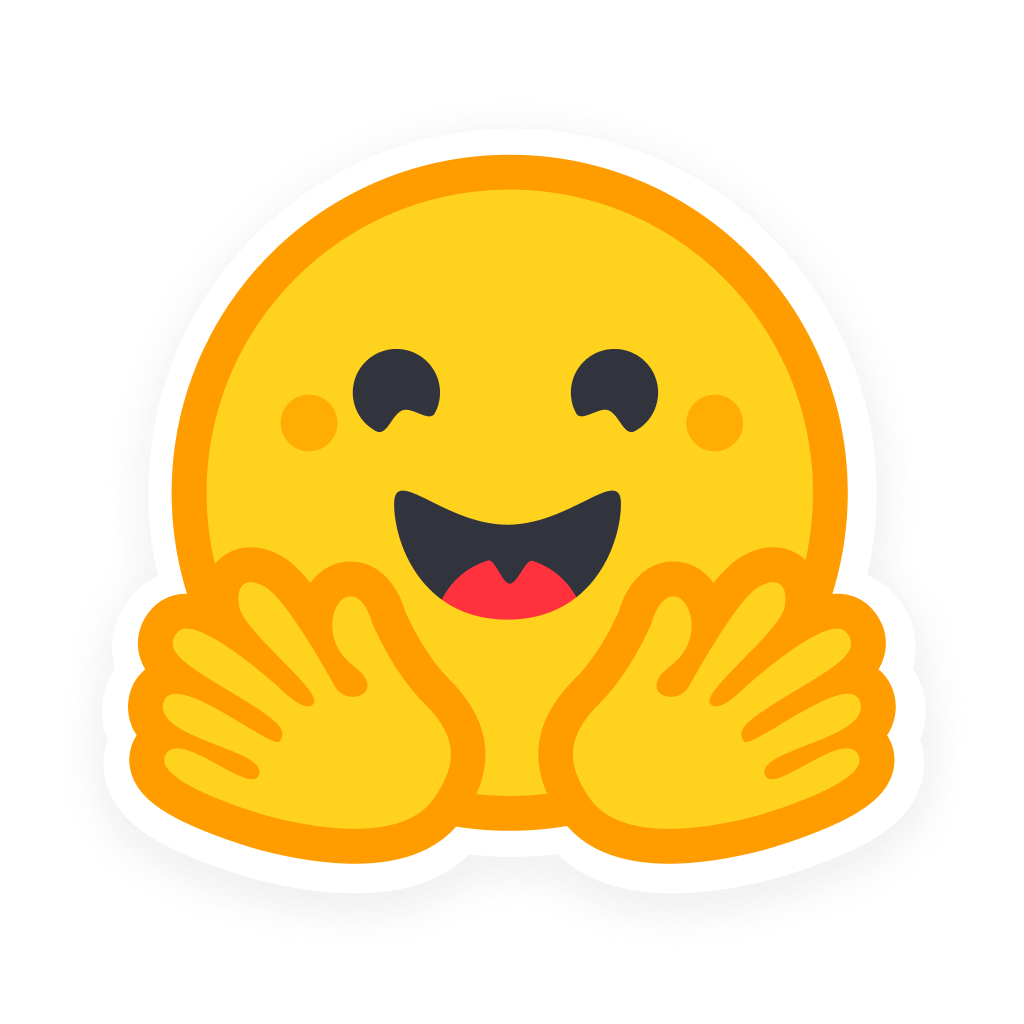}}}

\title{\textcolor{casepurple}{A*-Thought-V2}: Efficient Latent Reasoning \\ via Geometric Dynamics of LLM}

\author{%
Xiaoang Xu$^{1,5}$
~~Siyuan Liu$^{2,5}$
~~Shuo Wang$^{3}$
~~Junlan Feng$^{4}$
~~Fanyu Meng$^{4}$
~~Zhu Zhang$^{3,5}$\\
\bf Jixun Wang$^{1,5}$
~~Xiaorong Wang$^{5}$
~~Zihan Zhou$^{5}$
~~Xin Li$^{5}$
~~Chaojun Xiao$^{5}$\\
\bf Yiming Zhang$^{1}$
~~Huijia Wu$^{1}$%
\thanks{Corresponding author. Email: \texttt{xuxiaoang@bupt.edu.cn}.}
~~Liuyu Xiang$^{1}$
~~Peipei Li$^{1}$
~~Zhaofeng He$^{1}$\\[5pt]
\begin{tabular}{@{}l@{}}
$^{1}$Beijing University of Posts and Telecommunications\\
$^{2}$The Hong Kong Polytechnic University\quad
$^{3}$Tsinghua University\\
$^{4}$JIUTIAN Research, Beijing, China\quad
$^{5}$OpenBMB
\end{tabular}\\[15pt]
\github\ \textbf{GitHub:}\quad
\href{https://github.com/AI9Stars/AStar-Thought}
{\texttt{github.com/AI9Stars/AStar-Thought}}\\
\huggingface\ \textbf{Hugging Face:}\quad
\href{https://huggingface.co/datasets/xxang/AStar-Thought-V2-OpenR1-Math-3k}
{\texttt{AStar-Thought-V2-OpenR1-Math-3k}}
}

\begin{document}

\maketitle

\begin{abstract}
Chain-of-Thought (CoT) improves the reasoning ability of Large Language Models (LLMs) but incurs substantial computation and context costs. Existing methods either lose intermediate information through hard pruning or lack a principled criterion for continuous compression. We present \textbf{A*-Thought-V2}, a geometric dynamics of LLM guided framework that models CoT as a hidden-state trajectory and replaces hard deletion with an explicit--implicit interleaved latent architecture. After projecting question, step, and solution representations into a 3D PCA space, it measures alignment between each local transition and global question-to-solution direction. Aligned steps remain explicit text, whereas deviating steps are compressed into continuous latent tokens. Directional angles capture both local semantics and reasoning dynamics: small angles indicate direct execution and answer formation, while large angles more frequently involve checking, correction, and branch exploration; their temporal variation reveals exploration, convergence, and refinement stages. To train this architecture, we introduce stepwise embedding forcing, which pools each redundant step into a single latent embedding, and label forcing, which supervises that latent token with a soft multi-modal vocabulary distribution instead of a hard one-hot label. Experiments on Qwen3.5-9B and Qwen3.6-27B across six in-domain and out-of-domain benchmarks show that A*-Thought-V2 improves average accuracy by up to 2.6\% while reducing response length by up to half, increasing Accuracy per Computation Unit by 2.29$\times$, and reducing preprocessing and training time by 94.6\% and up to 80.3\%, respectively. Representation analyses suggest that latent states form a compact region distinct from textual states, while higher entropy at latent-token positions reflects broader soft targets that encourage richer step-level feature learning.
\end{abstract}

\begin{figure}[ht]
  \centering
    \subfigure[Chain-of-Thought]{
		\includegraphics[width=0.32\linewidth]{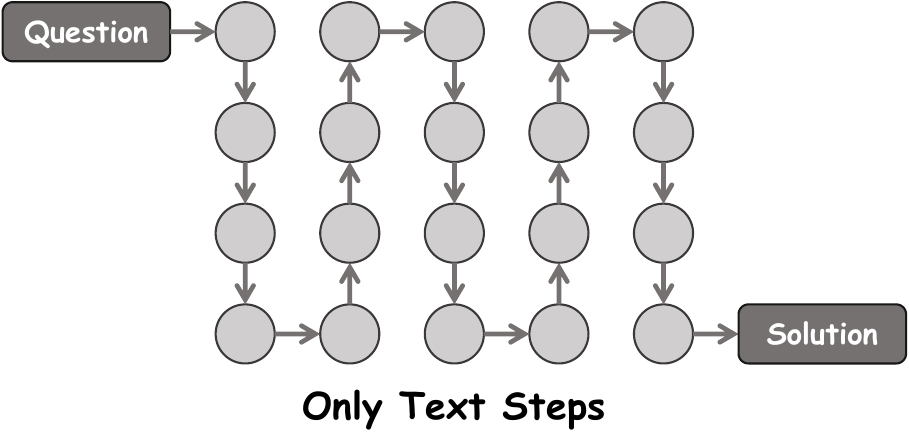}}
	\subfigure[A*-Thought]{
		\includegraphics[width=0.32\linewidth]{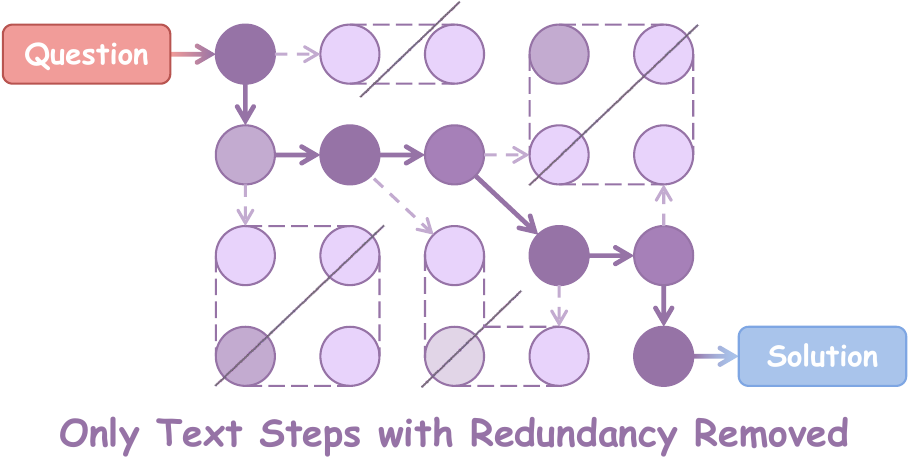}}
	\subfigure[A*-Thought-V2]{
		\includegraphics[width=0.32\linewidth]{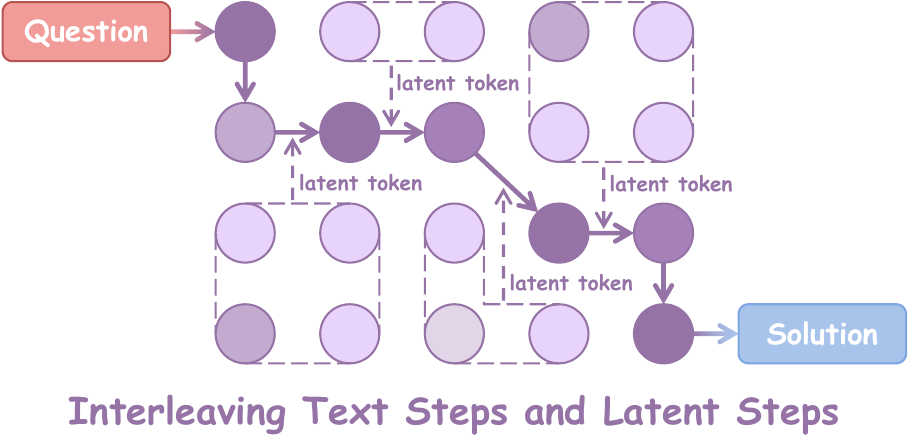}}
  \caption{Comparison of reasoning paradigms. A*-Thought applies hard pruning, retaining selected steps and discarding others. A*-Thought-V2 encodes redundant steps as latent representations, forming an information-dense explicit--implicit interleaved sequence that preserves reasoning progression.}
  \label{fig:intro}
\end{figure}

\section{Introduction}

\begin{figure}[ht]
  \centering
  \includegraphics[width=0.9\linewidth]{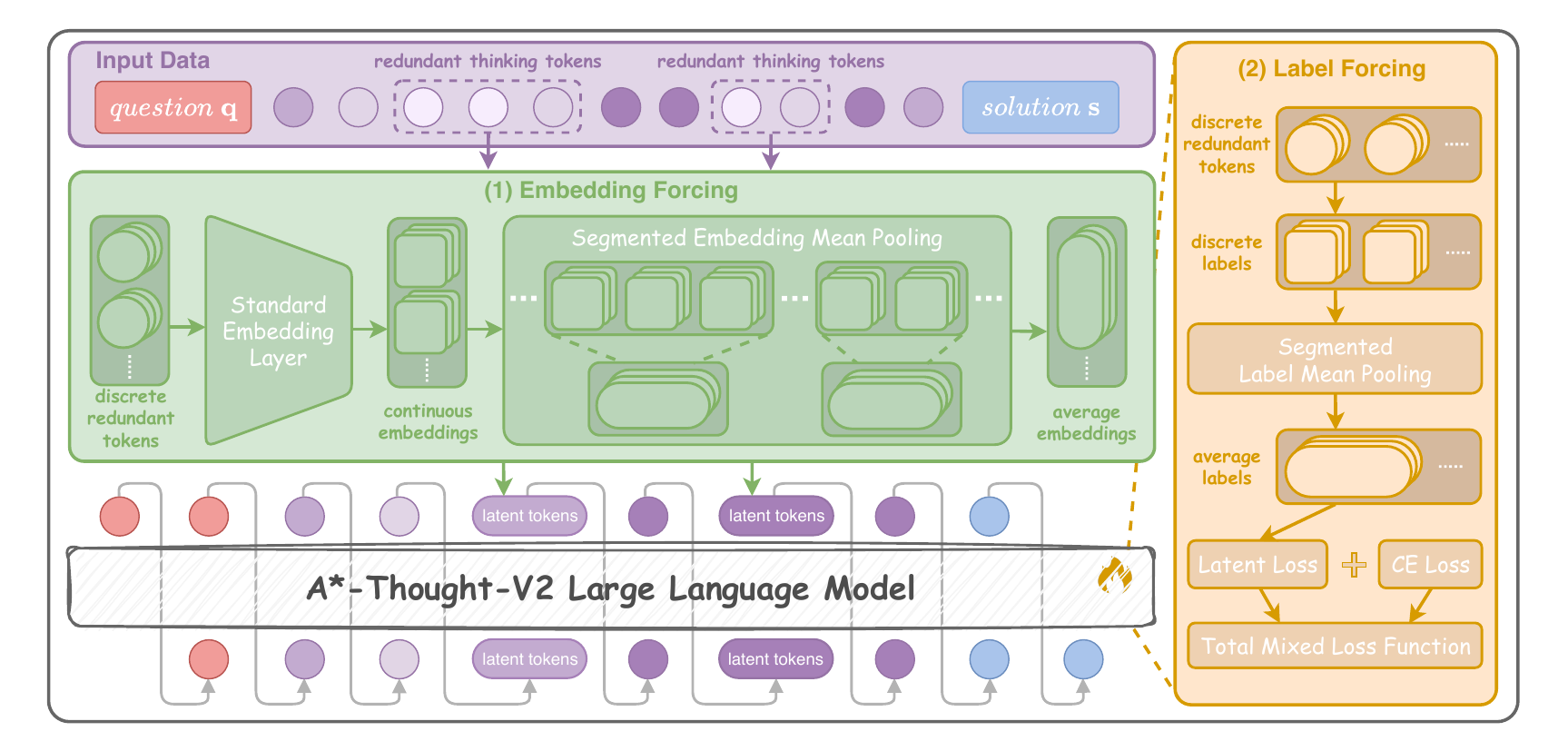}
  \caption{Framework of A*-Thought-V2. By heuristically constructing an explicit-implicit interleaved sequence, it compresses discrete redundant thinking tokens into continuous latent tokens via \textit{Embedding Forcing} and supervises them via \textit{Label Forcing}, integrating the latent loss with the standard cross-entropy loss.}
  \label{fig:framework}
\end{figure}

Chain-of-Thought (CoT) enhances LLM reasoning but incurs computational and context costs~\cite{s1, Chen2026, Liu2026}. Existing compression methods adopt hard compression, retaining selected steps while discarding others~\cite{A*-Thought,recut,tokenskip}, which may remove useful reasoning information. Latent reasoning instead encodes compressed steps into dense representations~\cite{PCCOT,ConceptLM,ponderlm}, but lacks an effective mechanism for selecting explicit steps and preserving the remaining information.

In this paper, we introduce A*-Thought-V2, a dynamics-guided framework for information-preserving CoT compression with an explicit--implicit interleaved architecture. Unlike A*-Thought~\cite{A*-Thought}, which directly discards redundant steps, A*-Thought-V2 models reasoning as a trajectory in the LLM hidden-state space and encodes redundant steps in the latent space. We project the question, intermediate steps, and solution into a three-dimensional PCA space and measure the alignment between each local transition and the global question-to-solution direction. Our analysis identifies six $30^\circ$ directional-angle intervals with semantic tendencies ranging from direct execution to correction and branch reconsideration. The temporal angle variation further characterizes three qualitative reasoning stages: exploration, convergence, and refinement. These geometric dynamics guide controllable explicit--latent step allocation and compression.

To learn the resulting interleaved sequence, we develop an SFT paradigm with \textit{Embedding Forcing} and \textit{Label Forcing}, as illustrated in Figure~\ref{fig:framework}. Embedding Forcing converts compressed spans into segment-pooled latent embeddings, while Label Forcing supervises the corresponding latent positions with soft vocabulary distributions. Joint optimization of the latent and standard text losses encourages compact latent tokens to preserve the semantics of compressed reasoning.

The main contributions of A*-Thought-V2 are summarized as follows:
\begin{itemize}
\item We provide a geometric dynamics analysis of CoT trajectories that identifies six directional-angle intervals with distinct semantic tendencies and three reasoning stages, thereby guiding controllable and efficient data selection and compression.
\item We develop an explicit--implicit interleaved latent architecture with step-wise embedding forcing and label forcing, replacing hard pruning with continuous latent representations supervised by step-level soft vocabulary distributions.
\item Extensive experiments across two model scales and six benchmarks show that A*-Thought-V2 improves average accuracy by up to 2.6\% and ACU by 2.29$\times$ over the baseline, while reducing compression time by 94.6\% over A*-Thought. Ablation and representation analyses further validate the proposed latent mechanisms.
\end{itemize}

\section{Preliminaries}

\begin{figure}[ht]
    \centering
    \subfigure[CoT Trajectory]{
        \includegraphics[width=0.30\linewidth]{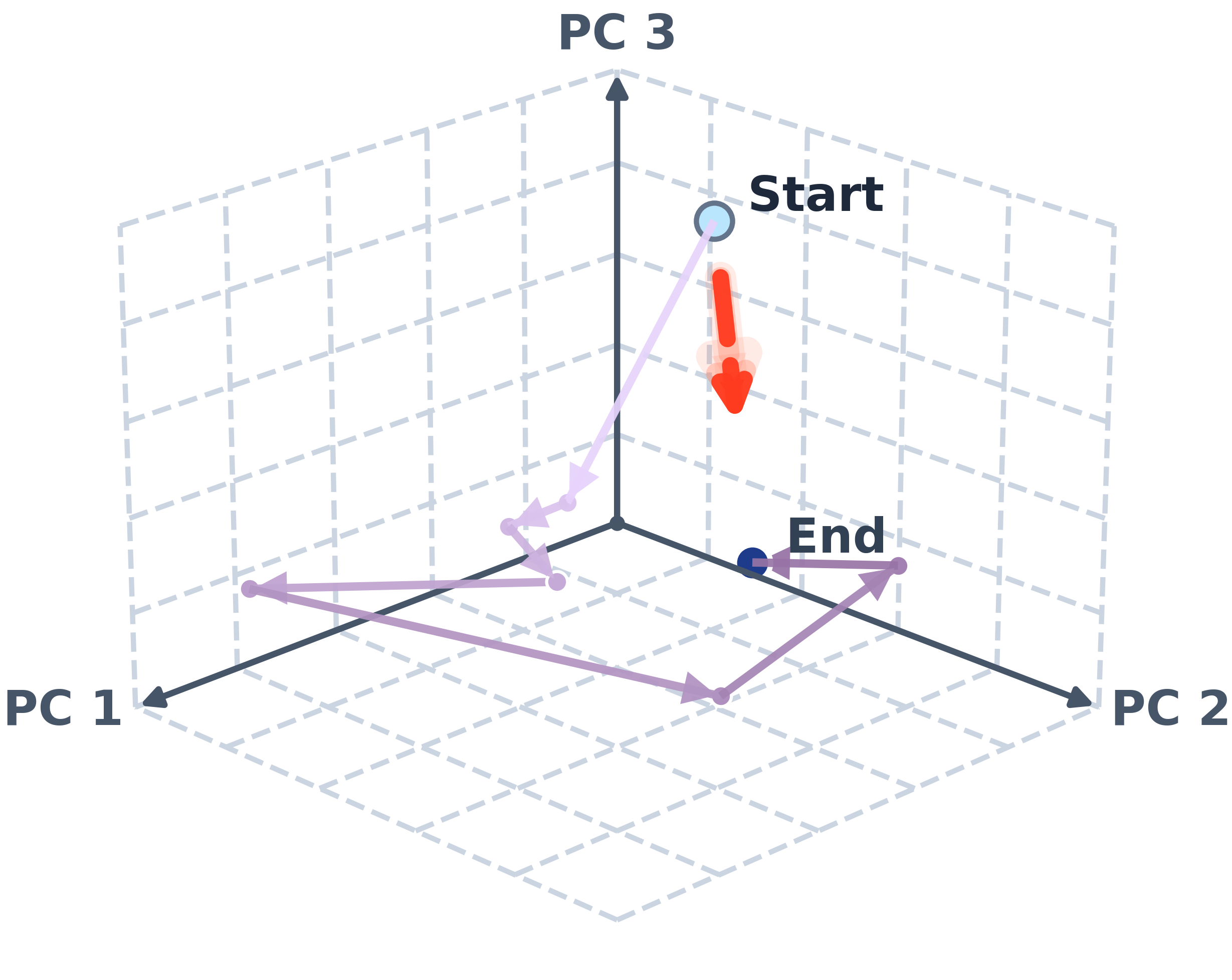}}
    \subfigure[Step Semantics]{
        \includegraphics[width=0.24\linewidth]{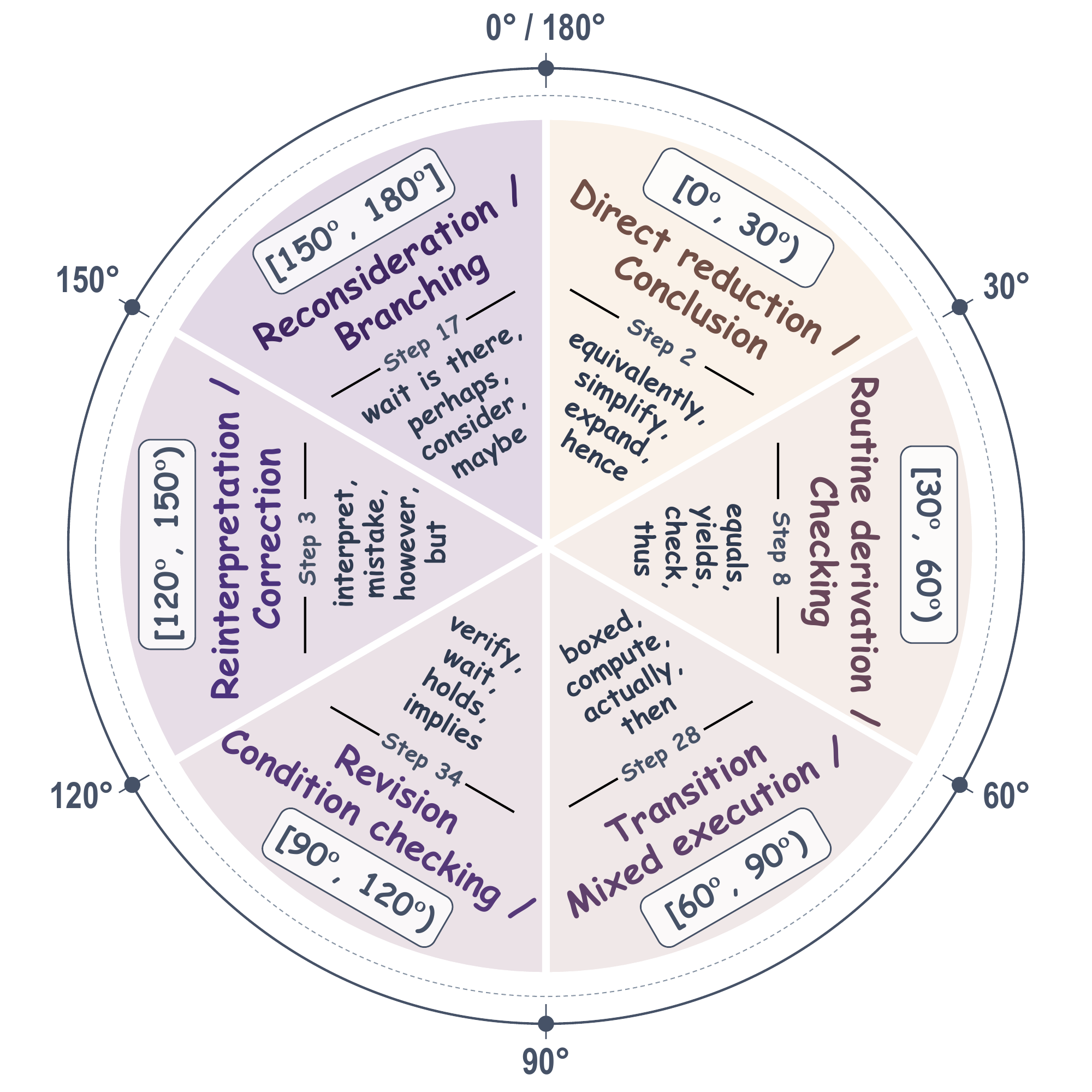}}
    \subfigure[Reasoning Stages]{
        \includegraphics[width=0.40\linewidth]{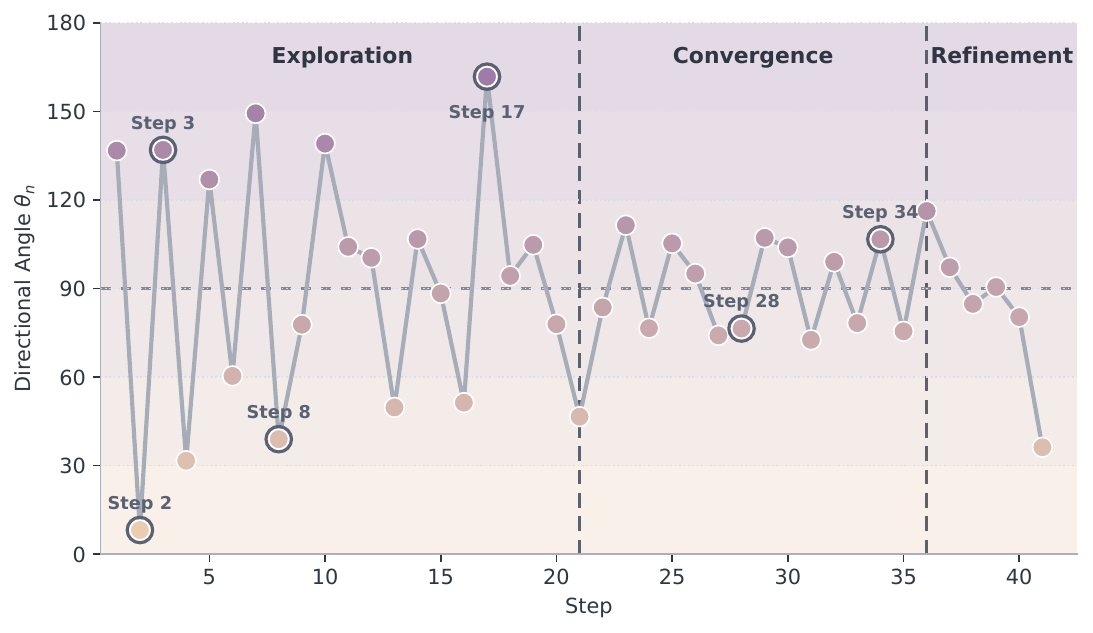}}
    \caption{PCA-projected CoT trajectory and its directional dynamics. (a) Three-dimensional representation trajectory. (b) Semantic tendencies across six angle intervals. (c) Exploration, convergence, and refinement stages reflected by step-wise angle variations.}
    \label{fig:pca}
\end{figure}

For a $T$-step CoT, most existing compression methods adopt hard compression, preserving steps with $\mathrm{flag}^{(n)}=1$ while directly discarding those with $\mathrm{flag}^{(n)}=0$. A*-Thought~\cite{A*-Thought} searches for this binary flag sequence in a two-dimensional tree structure. In contrast, A*-Thought-V2 encodes steps with $\mathrm{flag}^{(n)}=0$ into latent representations to retain more reasoning information at higher density, and models the reasoning process $\mathbf{q}\rightarrow\mathbf{t}^{(1)}\rightarrow\cdots\rightarrow\mathbf{t}^{(T)}\rightarrow\mathbf{s}$ as a trajectory in the PCA-projected hidden-state space, as illustrated in Figure~\ref{fig:pca}(a). After projecting the hidden states into three-dimensional space using PCA, we denote the representations of the question, the $n$-th reasoning step, and the solution as $\mathbf{h}_q$, $\mathbf{h}_{t^{(n)}}$, and $\mathbf{h}_s$, respectively. We define the global question-to-solution direction as $\mathbf{z}_0=\mathbf{h}_s-\mathbf{h}_q$, and the local transition directions as $\mathbf{z}_1=\mathbf{h}_{t^{(1)}}-\mathbf{h}_q$ and $\mathbf{z}_n=\mathbf{h}_{t^{(n)}}-\mathbf{h}_{t^{(n-1)}}$ for $n=2,\ldots,T$. We then quantify the directional alignment between each local transition and the global solution direction using their included angle:
\begin{equation}
    \theta_n=\arccos\left(
    \frac{\mathbf{z}_n^\top\mathbf{z}_0}
    {\|\mathbf{z}_n\|\|\mathbf{z}_0\|}
    \right).
\end{equation}
We analyze the PCA-projected CoT trajectory from both semantic and temporal perspectives. As shown in Figure~\ref{fig:pca}(b), small-angle transitions are mainly associated with direct reduction and answer formation, whereas large-angle transitions more frequently involve checking, correction, reinterpretation, and branch reconsideration. Detailed interval proportions, semantic categories, and lexical markers are provided in Appendix~\ref{app:angle-semantic-analysis}. Figure~\ref{fig:pca}(c) further reveals three reasoning stages from the temporal variation of $\theta_n$: exploration with strong oscillations, convergence with decreasing variation, and refinement before the final answer. The complete reasoning content corresponding to the marked steps is presented in Appendix~\ref{app:step-dynamics-case}.

\section{Methodology}

In this section, we first introduce the latent architecture construction method of A*-Thought-V2, Secondly, we present the modifications made during inference, with the aim of maintaining strict consistency between training and inference.

\subsection{Latent Architecture}

To enable LLMs to compress and learn high-dimensional latent thinking representations, we model latent tokens by embedding forcing and label forcing.

\paragraph{Embedding Forcing}

In a standard LLM, a sequence of discrete input tokens $\mathbf{x} = (\mathbf{x}_{1}, \mathbf{x}_{2}, \dots, \mathbf{x}_{n})$ is mapped by the embedding layer into continuous dense vector representations. At the token-level, the embedding of $i$-th token is:
\begin{equation}
\mathbf{e}_{i} = W_E(\mathbf{x}_{i}) \in \mathbb{R}^d
\end{equation}
where $W_E \in \mathbb{R}^{V \times d}$ is a learnable embedding matrix, $V$ is the vocabulary and $d$ is the hidden dimension. Then $\mathbf{e} = (\mathbf{e}_{1}, \mathbf{e}_{2}, \dots, \mathbf{e}_{n})$ proceeds to the subsequent transformer block, resulting in the corresponding hidden states $\mathbf{h} = (\mathbf{h}_{1}, \mathbf{h}_{2}, \dots, \mathbf{h}_{n})$.

In A*-Thought-V2, to achieve both less information loss and higher information compression density, we map redundant thinking steps $\{ \mathbf{t}^{(n)} \in \mathbf{t} \mid \mathrm{flag}^{(n)}=0 \}$ into the latent space via embedding forcing, transforming the previously hard-pruned steps into continuous latent thought vectors. For a redundant thinking step $\mathbf{t}^{(n)}$ consisting of $l^{(n)}$ discrete tokens, let its corresponding variable-length standard embedding sequence be $\mathbf{e}^{(n)} = (\mathbf{e}_1, \dots, \mathbf{e}_{l^{(n)}})$. To preserve the chronological logic and alleviate the information bottleneck, we map this step-level embedding sequence into a single latent token through segmented mean pooling:
\begin{equation}
\mathbf{c}^{(n)} = \frac{1}{l^{(n)}} \sum_{j=1}^{l^{(n)}} \mathbf{e}_{j}^{(n)} \in \mathbb{R}^{d},
\end{equation}
where $\mathbf{c}^{(n)}$ is the step-level latent embedding corresponding to the step $\mathbf{t}^{(n)}$. At the span-level, the step-level latent embeddings corresponding to a redundant span of $m$ consecutive reasoning steps are concatenated to form a latent span:
\begin{equation}
\mathbf{c} = \left( \mathbf{c}^{(1)}; \mathbf{c}^{(2)}; \dots; \mathbf{c}^{(m)} \right) \in \mathbb{R}^{m \times d}.
\end{equation}
The resulting latent span is then interleaved into the reasoning path $\mathbf{t}$ in place of the original embeddings of the redundant text tokens:
\begin{equation}
\mathbf{e}_{\mathbf{t}^{\prime}} = 
( \mathbf{e}_{1}; \mathbf{e}_{2}; \dots; \mathbf{c}_{1}; \dots; \mathbf{e}_{i}; \mathbf{e}_{i+1}; \dots; \mathbf{c}_{2}; \dots; \mathbf{e}_{n} ),
\end{equation}
where $\mathbf{e}_{\mathbf{t}^{\prime}}$ is text and latent tokens embeddings interleaved thinking path $\mathbf{t}^{\prime}$. During supervised training, the complete teacher-forced embedding sequence is formed by concatenating the question, interleaved reasoning trajectory, and solution representations:
\begin{equation}
\mathbf{e}_{input} = ( \mathbf{e}_{\mathbf{q}}; \mathbf{e}_{\mathbf{t}^{\prime}}; \mathbf{e}_{\mathbf{s}} ),
\end{equation}
where $\mathbf{e}_{\mathbf{q}}$ and $\mathbf{e}_{\mathbf{s}}$ denotes the standard embeddings of the question $\mathbf{q}$ and solution $\mathbf{s}$. Through this streamlined embedding forcing mechanism, A*-Thought-V2 effectively compresses lengthy redundant text into dense latent spatial representations. This allows the model to retain crucial heuristic reasoning information as a continuous semantic flow, while significantly reducing the overall context window occupancy.

\paragraph{Label Forcing}

Standard LLMs are trained through next-token prediction using deterministic one-hot targets. Let $\hat{p}_{i,v}$ denote the predicted probability of vocabulary item $v$ at token position $i$. For a sequence of length $N$, the standard cross-entropy loss is
\begin{equation}
\mathcal{L}_{CE}
=
-\frac{1}{N}
\sum_{i=1}^{N}
\sum_{v=1}^{V}
y_{i,v}\log\hat{p}_{i,v},
\end{equation}
where $y_{i,v}\in\{0,1\}$ is the one-hot target distribution.

In A*-Thought-V2, each latent token is expected to simultaneously encode the semantic information of multiple original tokens, requiring a multi-modal target distribution. Deterministic hard labels are therefore ill-suited for this objective. To supervise the compressed representations generated by embedding forcing, we introduce a step-wise label forcing mechanism using continuous soft probability distributions instead of discrete one-hot vectors.

Corresponding to the step-level compression strategy described in embedding forcing, we map all target tokens of a redundant step into a single multi-modal distribution. For the redundant step $\mathbf{t}^{(n)}$ consisting of $l^{(n)}$ tokens, the target soft label $\mathbf{y}_{soft}^{(n)} \in \mathbb{R}^V$ is constructed by averaging the one-hot encoded ground truth vectors of all tokens within that entire step:
\begin{equation}
\mathbf{y}_{soft}^{(n)} = \frac{1}{l^{(n)}} \sum_{j=1}^{l^{(n)}} \mathbf{y}^{(n)}_{j}.
\end{equation}
This multi-modal probability distribution acts as a segment-wise soft vocabulary target, forcing the latent token to simultaneously predict the macro-semantics of its corresponding reasoning step. The cross-entropy loss for this specific latent token, denoted as $\mathcal{L}_{latent}^{(n)}$, is then calculated between the latent output distribution and the step-wise soft target:
\begin{equation}
\mathcal{L}_{latent}^{(n)} = - \sum_{v=1}^{V} \mathbf{y}_{soft, v}^{(n)} \log \hat{p}_{n,v},
\end{equation}
where $\hat{p}_{n,v}$ is the model's predicted probability for the $v$-th vocabulary item at the latent position corresponding to step $\mathbf{t}^{(n)}$.

Finally, the overall training objective integrates the standard hard-label loss for regular text tokens and the soft-label loss for latent tokens. To ensure the model adequately learns this dense conceptual compression and strictly adheres to the latent structural boundaries, we apply a structural scaling weight $\lambda$ to the latent loss. The final mixed loss is formulated as:
\begin{equation}
\mathcal{L} = \frac{1}{N_{valid}} \left( \sum_{i \in \mathbb{I}_{text}} \mathcal{L}_{CE}^{(i)} + \lambda \sum_{k \in \mathbb{I}_{latent}} \mathcal{L}_{latent}^{(k)} \right),
\end{equation}
where $\mathbb{I}_{text}$ and $\mathbb{I}_{latent}$ denote the global index sets for text and latent tokens in the concatenated sequence respectively, and $N_{valid} = |\mathbb{I}_{text}| + |\mathbb{I}_{latent}|$ is the total number of valid tokens. Through label forcing, we explicitly supervise the internal logic alignment, guiding the model to match its deep latent representations with the chronological semantic distribution of the heuristically pruned steps.

\subsection{Inference}
\label{subsec:inference}

Each latent position is indexed by a discrete placeholder in the generated sequence but carries a continuous internal state in the same $d$-dimensional space as the hidden states $\mathbf{h}$. The placeholder and the begin and end of latent tags provide serialization and mode control; at latent positions, and the standard lookup embedding is replaced by the continuous state.

\paragraph{Latent Sampling}
After the begin of latent tag is generated at position $i$, the complete preceding context is retained in the autoregressive KV cache $\mathcal{C}_i$. At the first latent position, the last-layer hidden state of the boundary tag is used as the new input vector, while causal attention still accesses all preceding positions through the cache:
\begin{equation}
\mathbf{u}_{i+j}=\mathbf{h}_{i+j-1}^{(L)}, \qquad
\left(\mathbf{h}_{i+j}^{(L)},\mathcal{C}_{i+j}\right)
=\mathcal{T}\!\left(\mathbf{u}_{i+j};\mathcal{C}_{i+j-1}\right),
\qquad j=1,\ldots,m,
\end{equation}
where $\mathbf{u}_{i+j}$ is the input vector of the current latent position and $\mathcal{C}_{i+j-1}$ contains the keys and values of the entire preceding context. During training, $\mathbf{u}_{i+j}$ is replaced by the pooled embedding $\mathbf{c}^{(n)}$ of the corresponding compressed step, whereas inference uses the preceding last layer hidden state. Label forcing supervises each training time latent position with its step-wise soft vocabulary target. After the latent span ends, the model resumes standard text generation.

\section{Experiments}

\subsection{Setup}

\paragraph{Backbones and Training Data}
To evaluate the scalability of the proposed method, we conduct experiments on two open-source reasoning backbones with different model sizes: Qwen3.5-9B\footnote{\href{https://huggingface.co/Qwen/Qwen3.5-9B}{Qwen/Qwen3.5-9B}}, and Qwen3.6-27B\footnote{\href{https://huggingface.co/Qwen/Qwen3.6-27B}{Qwen/Qwen3.6-27B}}. This setup allows us to examine whether the proposed method remains effective across different capacity regimes. For training-based variants, we use OpenR1-Math-3k\footnote{\href{https://huggingface.co/datasets/TeichAI/deepseek-v3.2-speciale-openr1-math-3k}{TeichAI/deepseek-v3.2-speciale-openr1-math-3k}} as the supervised reasoning data.

\paragraph{Benchmarks}
We evaluate models on both in-domain and out-of-domain benchmarks. The in-domain benchmarks include Math500~\citep{math500}, AIME 2024, AIME 2025, and AIME 2026~\citep{aime}; the out-of-domain benchmarks include ARC-Challenge~\citep{arc} and GPQA-Diamond~\citep{gpqa}. Model performance is evaluated using the following metrics:
\begin{itemize}
\item {\em Accuracy}: The proportion of model outputs that match the ground-truth answers, measuring the model's correctness.
\item {\em Length}: The average length, i.e., the number of generated tokens, of the model's response; longer responses typically incur higher inference costs.
\item {\em Accuracy per Computation Unit} (ACU)~\citep{cotValve}: $\mathrm{ACU}=100\times\mathrm{Accuracy}/\mathrm{Length}$, with accuracy expressed in percentage points. A larger value indicates a better performance and efficiency trade-off.
\end{itemize}

\paragraph{Baselines}
We compare our method against the following baselines:
\begin{itemize}
    \item {\em SwiReasoning}~\citep{swireasoning}: A training-free method designed to improve the reasoning behavior of LRMs by interleaving text and latent mode.
    \item {\em CopT}~\citep{copt}: A pipeline that uses continuous-space verifiers to refine draft answers through on-policy reflection and correction.
    \item {\em A*-Thought}~\citep{A*-Thought}: A search-guided efficient reasoning method that explores candidate reasoning trajectories through hard compression.
\end{itemize}

\paragraph{Training and Evaluation Details}
We use a unified training and inference protocol for all trainable variants, with full hyperparameters reported in Appendix~\ref{app:hyperparameters}. In brief, models are trained for 3 epochs on 8 NVIDIA A100 80GB GPUs. For inference, we use temperature 1.0, top-$p$ 0.95, and repeat each evaluation 4 times. All remaining method-specific hyperparameters are reported in Tables~\ref{tab:training_hyperparams} and~\ref{tab:inference_hyperparams}.

\begin{table}[t]
\centering
\setlength{\tabcolsep}{4pt}
\renewcommand{\arraystretch}{1.15}
\resizebox{\textwidth}{!}{%
\begin{tabular}{l *{14}{r} c}
\toprule
\multirow{3}{*}{\textbf{Methods}}
& \multicolumn{8}{c}{\textbf{In-domain}}
& \multicolumn{4}{c}{\textbf{Out-of-domain}}
& \multicolumn{2}{c}{\multirow{2}{*}{\textbf{Average}}}
& \multirow{3}{*}{\textbf{ACU}} \\
\cmidrule(lr){2-9}
\cmidrule(lr){10-13}
& \multicolumn{2}{c}{Math500}
& \multicolumn{2}{c}{AIME 2024}
& \multicolumn{2}{c}{AIME 2025}
& \multicolumn{2}{c}{AIME 2026}
& \multicolumn{2}{c}{ARC-Challenge}
& \multicolumn{2}{c}{GPQA-Diamond}
& \multicolumn{2}{c}{}
& \\
\cmidrule(lr){2-3}
\cmidrule(lr){4-5}
\cmidrule(lr){6-7}
\cmidrule(lr){8-9}
\cmidrule(lr){10-11}
\cmidrule(lr){12-13}
\cmidrule(lr){14-15}
& Acc. & \#\,Tok.
& Acc. & \#\,Tok.
& Acc. & \#\,Tok.
& Acc. & \#\,Tok.
& Acc. & \#\,Tok.
& Acc. & \#\,Tok.
& Acc. & \#\,Tok.
& \\
\midrule

\rowcolor{lightergray}
Qwen3.5-9B
& 86.6 & 18453.63
& 85.0 & 35022.07
& 78.3 & 40545.69
& 84.2 & 37163.95
& 94.4 & 9859.94
& 81.8 & 29100.24
& 85.1 & 28357.59
& 0.30 \\

Qwen3.5-9B w/ OpenR1-Math-3k
& 91.9 & 6918.05
& 90.8 & 21019.25
& 83.3 & 26583.55
& 87.5 & 25423.37
& \underline{97.0} & 491.71
& \textbf{83.5} & 16510.11
& 89.0 & 16157.67
& 0.55 \\

\quad + SwiReasoning
& \textbf{97.8} & 3392.95
& 90.0 & 25085.87
& 86.7 & 29413.63
& 90.0 & 27549.57
& 95.6 & 629.74
& 79.8 & 21099.01
& 90.0 & 17861.79
& 0.50 \\

\quad + CopT
& 95.0 & 3829.28
& \textbf{93.3} & 23420.33
& 86.7 & 28707.10
& 86.7 & 30565.07
& 94.5 & 964.13
& 79.3 & 20054.05
& 89.3 & 17923.33
& 0.50 \\

\quad + A*-Thought
& 89.9 & 6422.13
& 70.0 & \textbf{18132.64}
& 60.8 & 25152.91
& 73.3 & 23899.95
& 96.8 & \underline{468.63}
& 81.3 & \textbf{13030.06}
& 78.7 & 14517.72
& 0.54 \\

\rowcolor{rowVtwoSixty}
\quad + A*-Thought-V2 w/ $\tau=60^\circ$
& \underline{97.0} & \underline{2918.68}
& \underline{91.7} & \underline{19167.03}
& \underline{89.2} & \underline{23509.88}
& \underline{90.0} & \underline{21098.94}
& 96.5 & 471.94
& 81.6 & \underline{13891.47}
& \underline{91.0} & \underline{13509.66}
& \underline{0.67} \\

\rowcolor{rowVtwoNinety}
\quad + A*-Thought-V2 w/ $\tau=90^\circ$
& 96.9 & \textbf{2743.97}
& 90.8 & 20266.38
& \textbf{90.0} & \textbf{21765.58}
& \textbf{91.7} & \textbf{20025.40}
& \textbf{97.1} & \textbf{449.51}
& \underline{82.8} & 15521.67
& \textbf{91.5} & \textbf{13462.08}
& \textbf{0.68} \\

\midrule

\rowcolor{lightergray}
Qwen3.6-27B
& 70.9 & 13911.97
& 84.2 & 29521.38
& 71.7 & 38445.60
& 77.5 & 36186.72
& 96.8 & 1157.41
& \underline{85.5} & 21114.27
& 81.1 & 23389.56
& 0.35 \\

Qwen3.6-27B w/ OpenR1-Math-3k
& 91.9 & 6926.81
& 95.8 & 16650.70
& \underline{94.2} & 21203.67
& 92.5 & 25769.07
& \textbf{97.4} & 466.43
& 85.5 & 13126.60
& 92.9 & 14023.88
& 0.66 \\

\quad + SwiReasoning
& \textbf{97.8} & 2679.25
& 90.0 & 19010.50
& 93.3 & 25873.40
& \textbf{96.7} & 25755.47
& 95.6 & 536.35
& 82.3 & 15469.61
& 92.6 & 14887.43
& 0.62 \\

\quad + CopT
& \underline{97.6} & 2979.45
& \underline{96.7} & 19024.27
& 86.7 & 26306.73
& 90.0 & 26089.37
& 95.1 & 943.54
& 84.9 & 15470.21
& 91.8 & 15135.60
& 0.61 \\

\quad + A*-Thought
& 88.0 & 8685.41
& 76.7 & \underline{15864.80}
& 65.8 & \textbf{18373.82}
& 70.8 & \textbf{18907.27}
& 96.9 & 473.21
& 84.0 & \textbf{10149.79}
& 80.4 & \underline{12075.72}
& 0.67 \\

\rowcolor{rowVtwoSixty}
\quad + A*-Thought-V2 w/ $\tau=60^\circ$
& 95.6 & \underline{2585.44}
& 95.9 & 17659.38
& \textbf{95.0} & 21320.58
& 91.7 & 20987.99
& 97.0 & \underline{463.19}
& 85.1 & 12405.54
& \underline{93.4} & 12570.35
& \underline{0.74} \\

\rowcolor{rowVtwoNinety}
\quad + A*-Thought-V2 w/ $\tau=90^\circ$
& 96.7 & \textbf{2496.34}
& \textbf{96.7} & \textbf{14931.38}
& 93.3 & \underline{20154.48}
& \underline{93.3} & \underline{20622.56}
& \underline{97.3} & \textbf{430.91}
& \textbf{86.0} & \underline{12077.05}
& \textbf{93.9} & \textbf{11785.45}
& \textbf{0.80} \\

\bottomrule
\end{tabular}%
}
\caption{Main results. The best results are shown in \textbf{bold}, and the second-best results are \underline{underlined}.}
\label{tab:main-results}
\end{table}

\subsection{Main Results}

The detailed experimental results in Table~\ref{tab:main-results} lead to the following key observations:

\paragraph{Up to 2.6\% accuracy gain and 2.29$\times$ ACU improvement.}
A*-Thought-V2 improves average accuracy by up to 2.6 percentage points over same-data SFT across both model scales. For comparison, the $90^\circ$ variant nearly halves response length and raises ACU from 0.35 to 0.80, corresponding to a 2.29$\times$ improvement over the original Qwen3.6-27B backbone. Similar gains are observed on Qwen3.5-9B.

\paragraph{Up to 16.0\% shorter responses than supervised fine-tuning with higher accuracy.}
Compared with SFT on OpenR1-Math-3k, A*-Thought-V2 produces more accurate and shorter reasoning traces. On Qwen3.6-27B, the $90^\circ$ variant improves average accuracy by 1.0 points while reducing response length by 16.0\%. It also surpasses SwiReasoning, CopT, and A*-Thought in average accuracy and ACU, showing the advantage of latent compression over training-free enhancement and hard pruning.

% \paragraph{Semantically grounded thresholds generalize across domains.}
% The gains span in-domain and out-of-domain reasoning benchmarks. Both thresholds follow the semantic partition of directional angles: small-angle steps remain explicit, while large-angle steps are compressed into latent tokens. The $90^\circ$ cut preserves forward-aligned execution and compresses checking, correction, and branch exploration; the stricter $60^\circ$ cut also compresses mixed execution and transition. Under both settings, A*-Thought-V2 improves accuracy over the same-data SFT baseline while shortening responses, with $90^\circ$ offering the strongest overall trade-off and $60^\circ$ remaining competitive. This consistency supports the semantic interpretation of the angle intervals.

\paragraph{Semantically grounded thresholds generalize across domains.} The $90^\circ$ threshold preserves forward-aligned steps while compressing checking, correction, and exploration into latent tokens; $60^\circ$ additionally compresses mixed execution and transition. Across the evaluated in-domain and out-of-domain benchmarks, both improve overall average accuracy and shorten responses relative to same-data SFT, with $90^\circ$ offering the best overall trade-off. These results are consistent with the proposed semantic partition.

\section{Analysis}

\subsection{Ablation on Step Selection and Latent Architecture}

\begin{table}[h]
\centering
\caption{Ablation study of geometric step selection, Embedding Forcing, and Label Forcing.}
\label{tab:ablation_ef_lf}
\setlength{\tabcolsep}{6pt}
\resizebox{\textwidth}{!}{%
\begin{tabular}{l|rrrrrr|rr|r}
\toprule
\multirow{2}{*}{\textbf{Methods}} & \multicolumn{2}{c}{\textbf{AIME 2024}} & \multicolumn{2}{c}{\textbf{AIME 2025}} & \multicolumn{2}{c}{\textbf{AIME 2026}} & \multicolumn{2}{|c|}{\textbf{Average}} & \multirow{2}{*}{\textbf{ACU}} \\
\cmidrule(lr){2-3} \cmidrule(lr){4-5} \cmidrule(lr){6-7} \cmidrule(lr){8-9}
 & \multicolumn{1}{c}{Acc. (\%)} & \multicolumn{1}{c}{\# Tokens} & \multicolumn{1}{c}{Acc. (\%)} & \multicolumn{1}{c}{\# Tokens} & \multicolumn{1}{c}{Acc. (\%)} & \multicolumn{1}{c}{\# Tokens} & \multicolumn{1}{|c}{Acc. (\%)} & \multicolumn{1}{c|}{\# Tokens} & \\
\midrule
A*-Thought-V2-Qwen3.6-27B w/ $\tau=90^\circ$ & \textbf{96.7}$_{\pm 0.0}$ & \textbf{14931.38}$_{\pm 802.13}$ & \textbf{93.3}$_{\pm 2.7}$ & 20154.48$_{\pm 993.78}$ & \textbf{93.3}$_{\pm 2.7}$ & 20622.56$_{\pm 1018.66}$ & \textbf{94.5}$_{\pm 0.9}$ & \textbf{18569.47}$_{\pm 449.14}$ & \textbf{0.51} \\
\midrule
\quad w/ Random Angles      & 94.2$_{\pm 2.7}$ & 16958.74$_{\pm 1374.64}$ & 90.8$_{\pm 4.2}$ & 22457.13$_{\pm 1534.80}$ & 93.3$_{\pm 1.7}$ & 20837.51$_{\pm 1063.29}$ & 92.8$_{\pm 1.7}$ & 20084.46$_{\pm772.86}$ & 0.46 \\
\quad w/ Reversed Selection & 93.3$_{\pm 2.7}$ & 18540.03$_{\pm 2055.31}$ & 90.8$_{\pm 1.7}$ & 22433.76$_{\pm 726.93}$ & 90.0$_{\pm 3.8}$ & 22990.10$_{\pm 1568.11}$ & 91.4$_{\pm 1.6}$ & 21321.30$_{\pm 895.15}$ & 0.43 \\
\midrule
\quad w/o Embedding Forcing & 94.2$_{\pm 3.3}$ & 18045.54$_{\pm 423.06}$ & 92.5$_{\pm 5.1}$ & 20259.82$_{\pm 1385.49}$ & 91.7$_{\pm 1.9}$ & \textbf{18888.45$_{\pm 1466.75}$} & 92.8$_{\pm 3.2}$ & 19064.60$_{\pm 655.20}$ & 0.49 \\
\quad w/o Label Forcing & 76.7$_{\pm 7.2}$ & 16514.35$_{\pm 1295.32}$ & 68.4$_{\pm 8.4}$ & \textbf{18763.83$_{\pm 2328.92}$} & 72.5$_{\pm 5.7}$ & 21032.20$_{\pm 1933.23}$ & 72.5$_{\pm 4.2}$ & 18770.13$_{\pm 1271.71}$ & 0.39 \\
\quad w/o Embedding Forcing \& Label Forcing & 59.2$_{\pm 7.4}$ & 16053.92$_{\pm 416.88}$ & 55.0$_{\pm 1.9}$ & 22145.06$_{\pm 1267.15}$ & 70.8$_{\pm 5.0}$ & 19437.83$_{\pm 1535.73}$ & 61.7$_{\pm 2.6}$ & 19212.27$_{\pm 56.01}$ & 0.32 \\
\bottomrule
\end{tabular}%
}
\end{table}

\paragraph{Effect of geometric step selection.} We compare geometry-guided selection with random-angle and reversed-selection variants, where the latter compresses small-angle steps and retains large-angle steps as text, The compression rates of the three are 48.09\%, 48.16\%, and 47.15\% respectively. Table~\ref{tab:ablation_ef_lf} shows that geometry-guided selection achieves higher average accuracy and ACU than both variants while reducing average generation length by 7.54\% and 12.91\%, respectively. These results support the chosen direction of explicit--latent allocation over the tested alternatives.

\paragraph{Roles of Embedding Forcing and Label Forcing.} Removing EF, LF, or both reduces average accuracy to 92.8\%, 72.5\%, and 61.7\%, respectively. These results highlight LF's importance and EF's complementary benefits, supporting their combined use for latent training.

\subsection{PCA Visualization of Latent Tokens}

\begin{figure}[ht]
    \centering
    \subfigcapskip=-3pt
    \subfigure[2D PCA]{
        \includegraphics[width=0.32\linewidth]{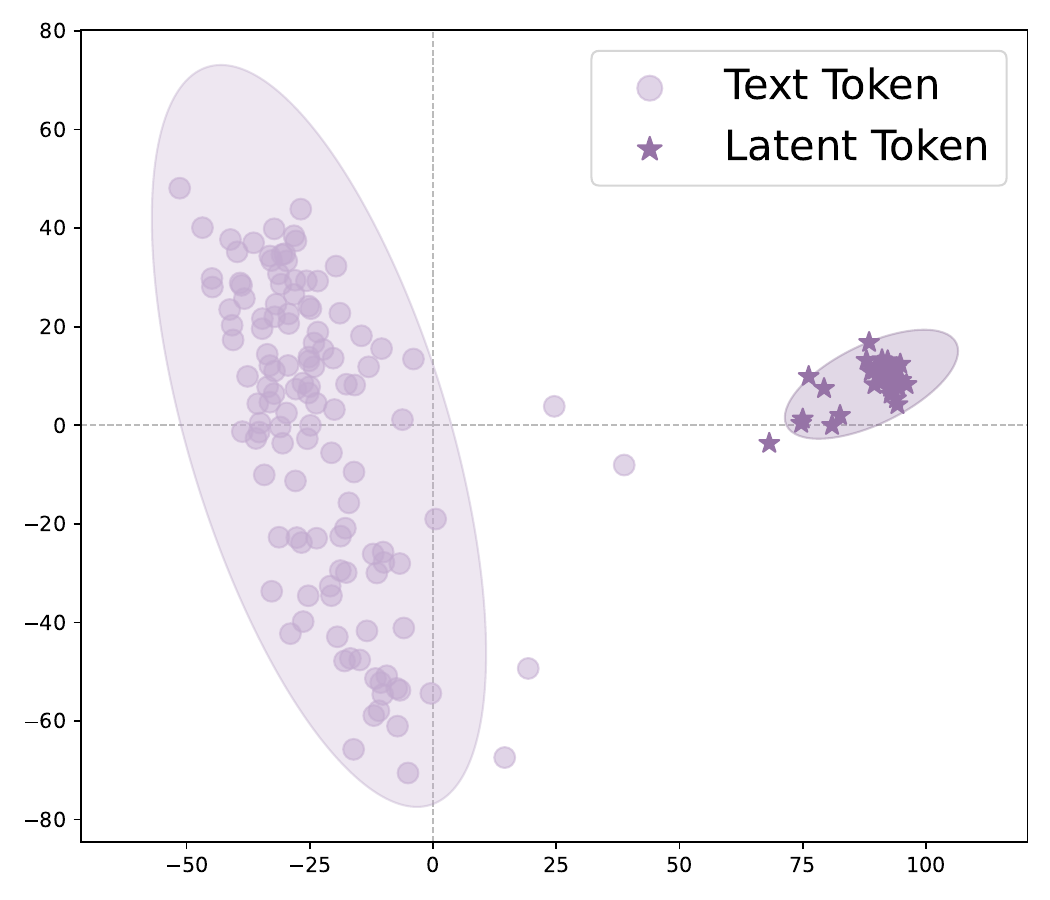}}
    \subfigure[3D PCA]{
        \includegraphics[width=0.40\linewidth]{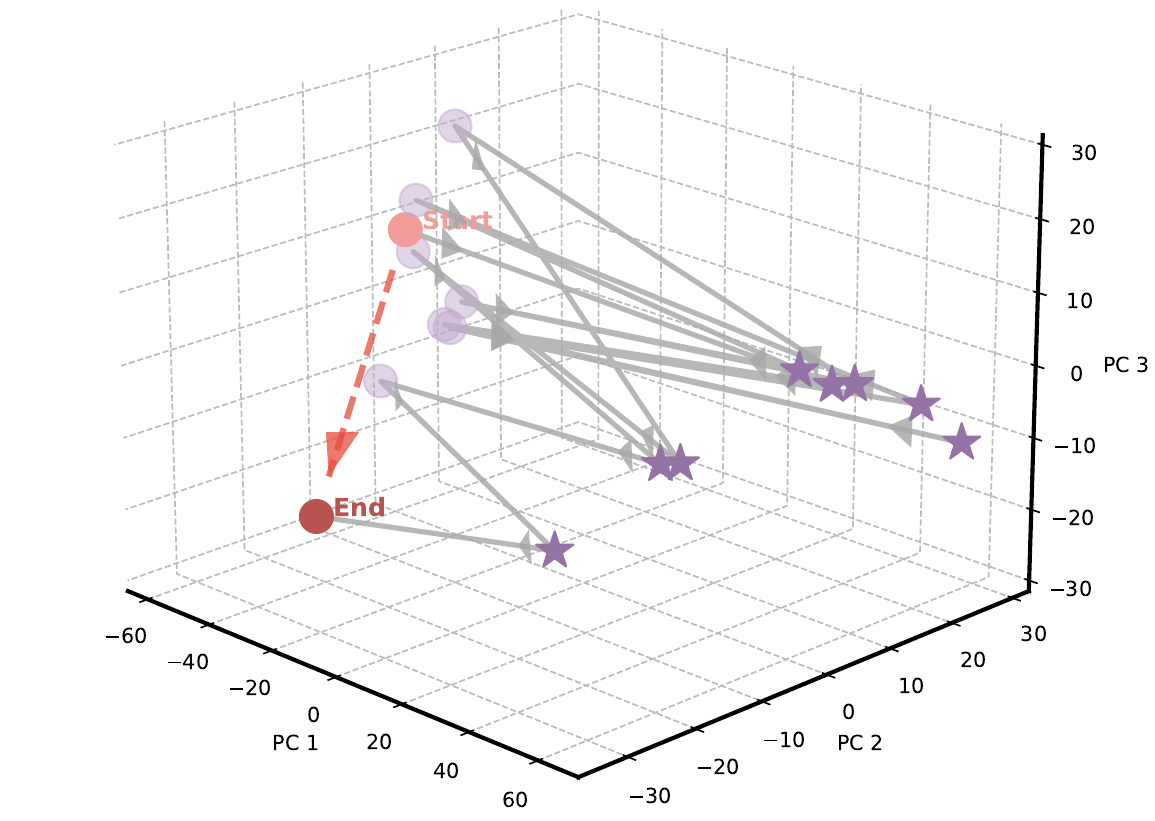}}
    \caption{PCA projections of text-token and latent-token hidden states produced by A*-Thought-V2-Qwen3.5-9B on AIME2024 case 79.}
    \label{fig:latent_pca_visualization}
\end{figure}

Figure~\ref{fig:latent_pca_visualization} visualizes the hidden states of text and latent tokens. In the 2D projection, latent tokens form a compact cluster distinct from the broader text-token distribution. The 3D projection shows that they occupy a coherent intermediate region along the reasoning trajectory. This suggests that latent tokens encode structured compressed reasoning information. Additional cases appear in Appendix~\ref{app:case}.

\subsection{Token Entropy Analysis under Label Forcing}

\begin{figure}[ht]
	\centering
	\subfigcapskip=-3pt
	\subfigure[Qwen3.5-9B]{
		\includegraphics[width=0.48\linewidth]{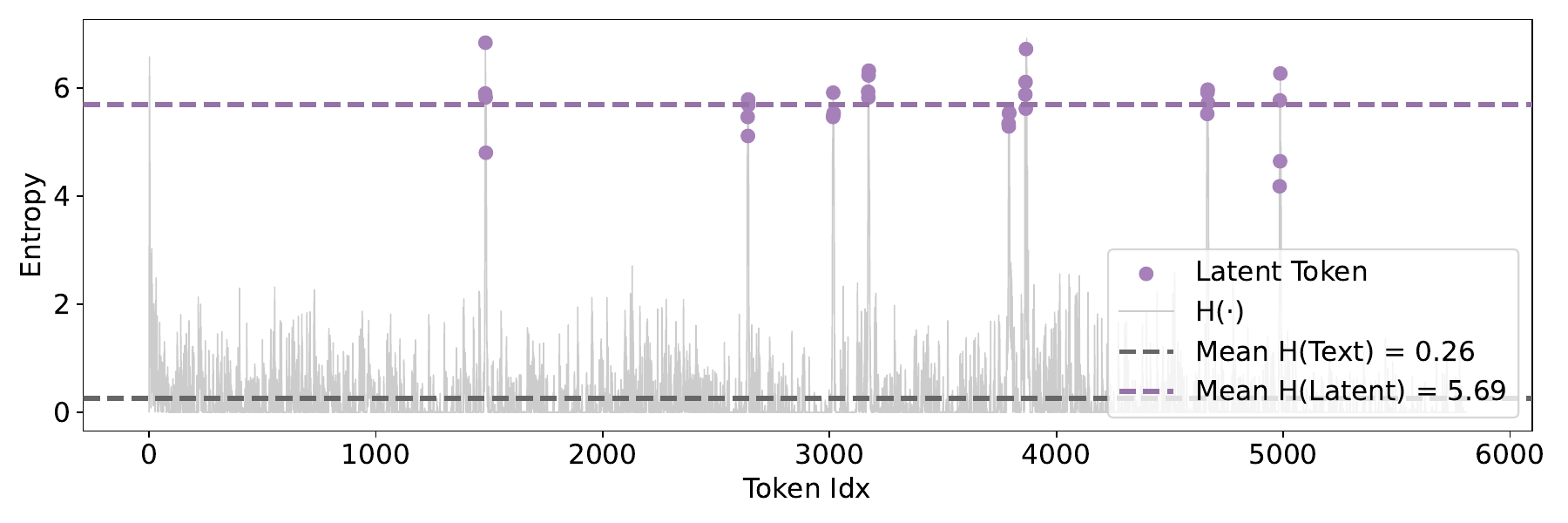}}
	\subfigure[Qwen3.6-27B]{
		\includegraphics[width=0.48\linewidth]{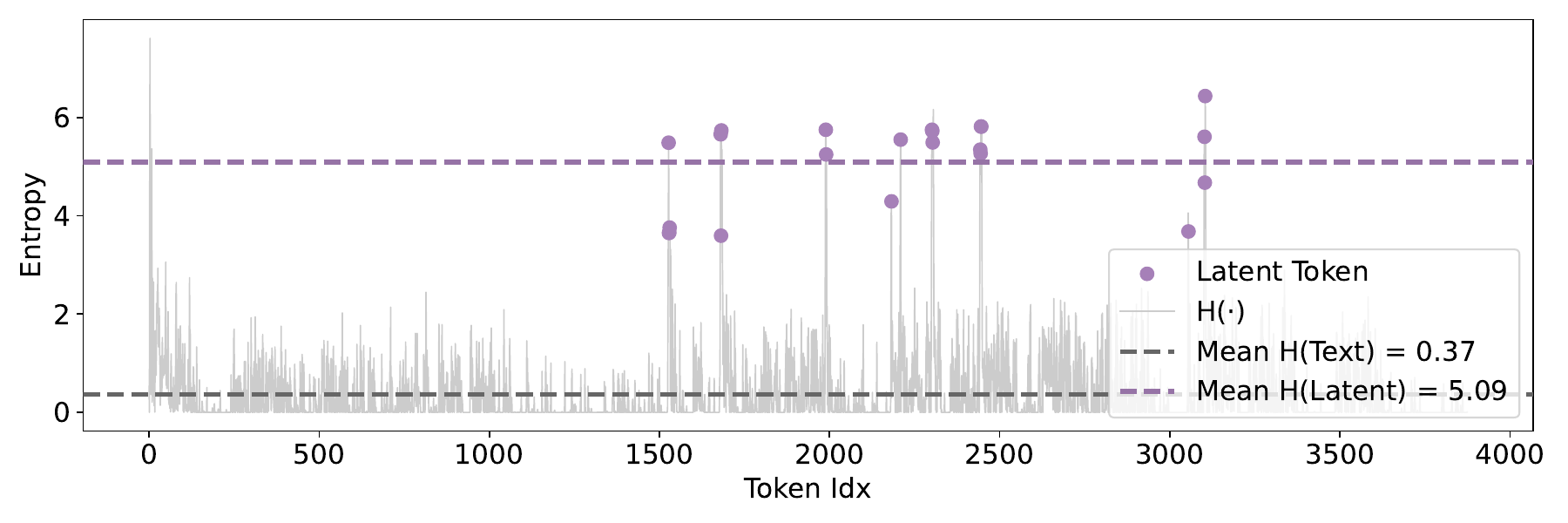}}
	\caption{The token entropy along the entire CoT trajectory.}
    \label{fig:entropy}
\end{figure}

Figure~\ref{fig:entropy} shows higher predictive entropy at latent-token than explicit-text positions across both model scales. Explicit tokens use one-hot targets, whereas latent tokens use soft vocabulary distributions aggregated from compressed-step tokens. Therefore, the higher predictive entropy primarily indicates that the model fits these broader soft targets.

\subsection{Effect of Maximum Latent Length}

\begin{figure}[H]
    \centering
    \subfigcapskip=-3pt
    \subfigure[Accuracy]{
        \includegraphics[height=0.19\linewidth]{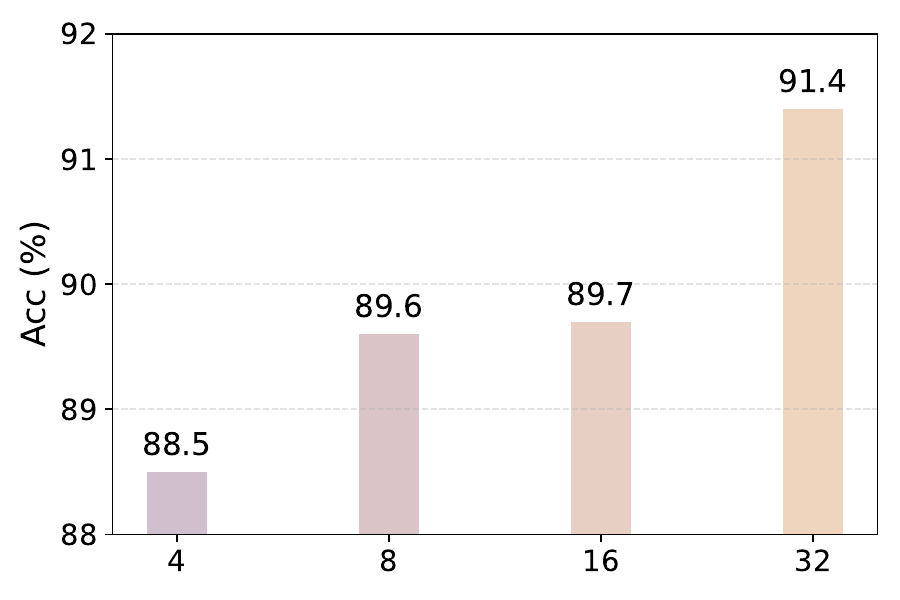}}%
    \subfigure[Length]{
        \includegraphics[height=0.19\linewidth]{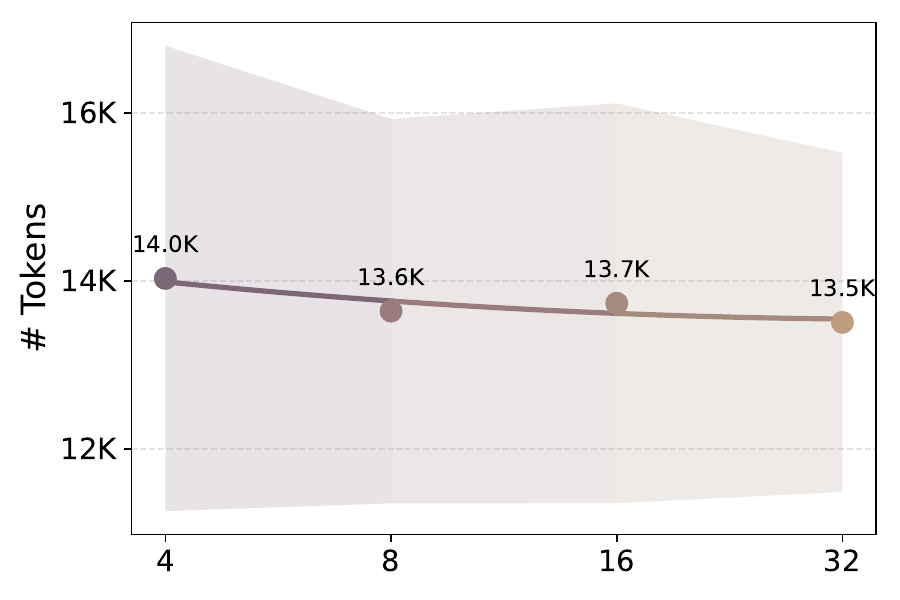}}%
    \subfigure[Latent Length Distribution]{
        \includegraphics[height=0.19\linewidth]{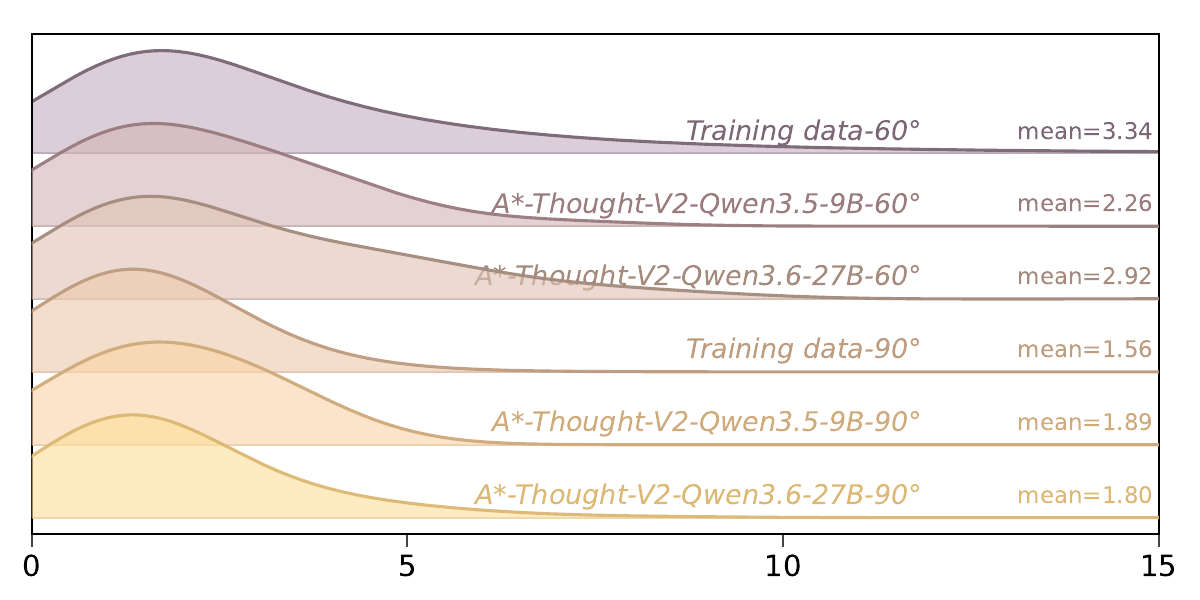}}
    \caption{Ablation study of max latent length on AIME2024 and GPQA-Diamond.}
    \label{fig:ablation_latent_len}
\end{figure}

We investigate the effect of the maximum latent length in Figure~\ref{fig:ablation_latent_len}. Figure~\ref{fig:ablation_latent_len}(a) and (b) show that a larger cap generally yields higher accuracy with shorter responses, suggesting that latent tokens compactly encode intermediate steps. Figure~\ref{fig:ablation_latent_len}(c) shows that the latent-length distribution of decoded outputs closely matches that of the training data, indicating that the model has effectively learned the latent reasoning format.

% The effect of maximum latent count is analyzed in Appendix~\ref{app:analysis_max_latent_count}.

\subsection{Training Dynamics}

\begin{figure}[ht]
    \centering
    \subfigcapskip=-3pt
    \subfigure[Qwen3.5-9B]{
        \includegraphics[width=0.44\linewidth]{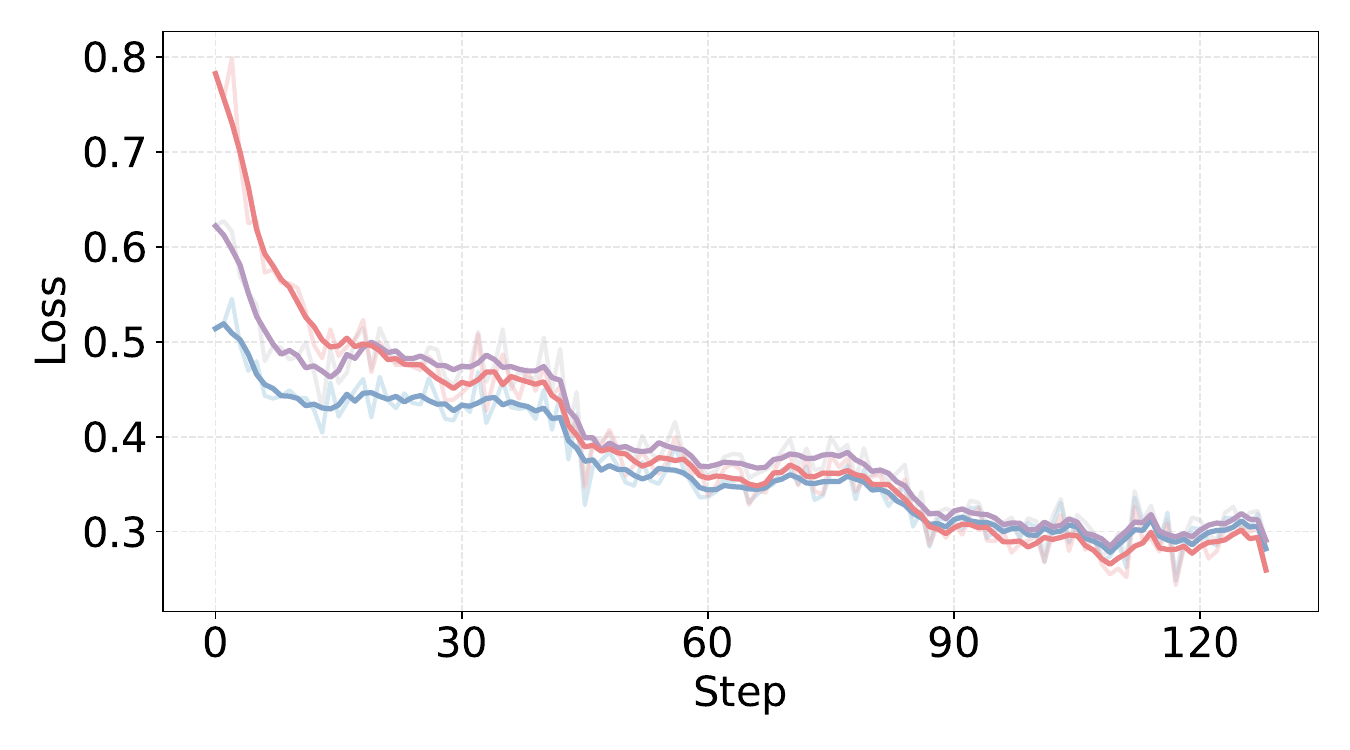}}
    \subfigure[Qwen3.6-27B]{
        \includegraphics[width=0.44\linewidth]{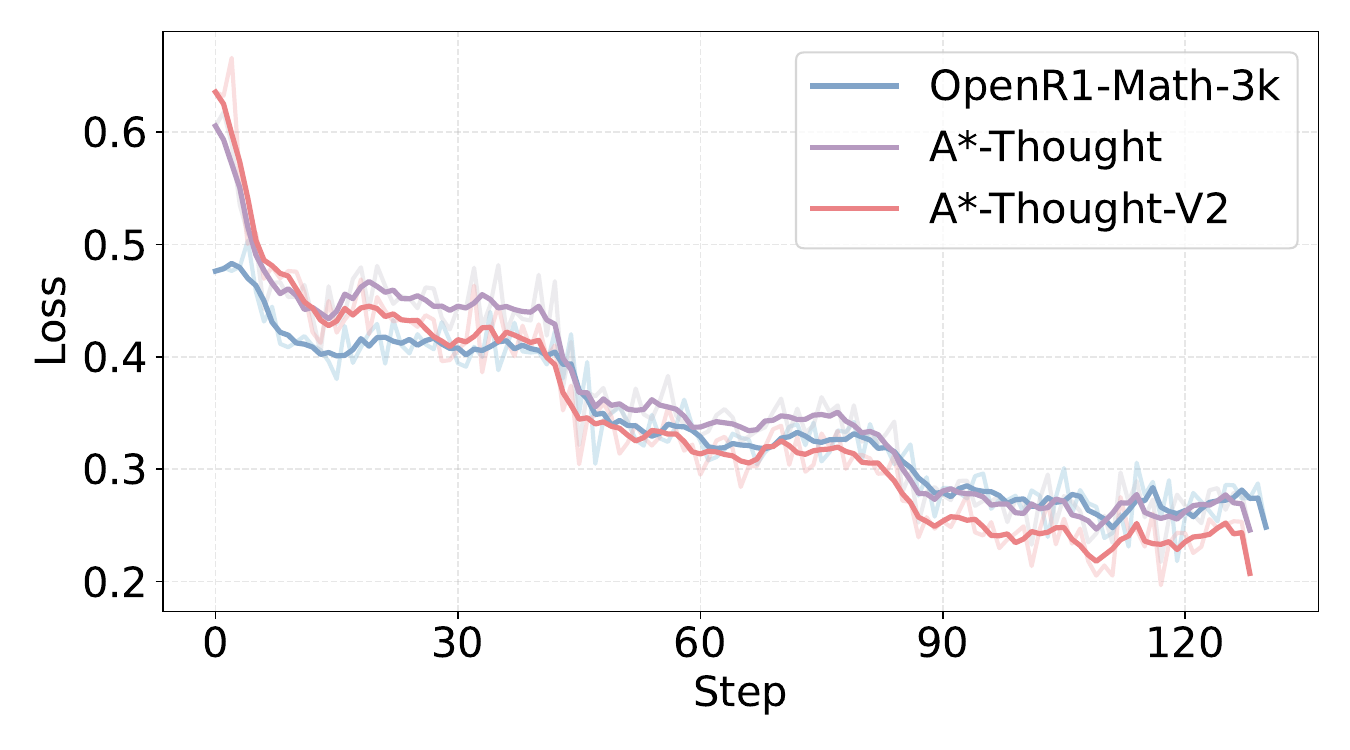}}
    \caption{Training-loss curves of different methods at two model scales.}
    \label{fig:loss}
\end{figure}

Figure~\ref{fig:loss} compares the training losses of OpenR1-Math-3k, A*-Thought, and A*-Thought-V2. Although A*-Thought-V2 begins with a higher loss due to learning latent reasoning segments, it converges rapidly and achieves the lowest final loss at both model scales. This suggests that the proposed solution-oriented latent compression not only preserves the learnability of the reasoning process, but also provides a more effective training signal after convergence.

\subsection{Compression and Training Efficiency}

Table~\ref{tab:compression_results} shows that A*-Thought-V2 reduces compression time from 5:16:22 to 0:16:57, a 94.6\% reduction over A*-Thought. The $60^\circ$ variant achieves the strongest compression rate (31.67\%) and reduces training time by 80.3\% and 68.7\% for Qwen3.5-9B and Qwen3.6-27B, respectively. The $90^\circ$ variant retains more reasoning tokens while still providing substantial training acceleration. These results demonstrate the efficiency and controllability of A*-Thought-V2 across model scales.

\begin{table}[ht]
\centering
\caption{Compression rates and preprocessing and training times across methods.}
\label{tab:compression_results}
\setlength{\tabcolsep}{3.5pt}
\resizebox{\textwidth}{!}{%
\begin{tabular}{l c c c c}
\toprule
\multirow{2}{*}{\textbf{Methods}}
& \textbf{Compression}
& \textbf{Compression}
& \multicolumn{2}{c}{\textbf{Train Time}} \\
\cmidrule(lr){4-5}
& \textbf{Rate}
& \textbf{Time}
& \textbf{Qwen3.5-9B}
& \textbf{Qwen3.6-27B} \\
\midrule

\rowcolor{lightergray}
OpenR1-Math-3k
& 100.00\%
& --
& 12:21:18
& 21:24:52 \\

~~~~+ A*-Thought
& 75.12\%
& 5:16:22
& 5:34:40 \textcolor{green}{\scriptsize (-54.9\%)}
& 10:26:22 \textcolor{green}{\scriptsize (-51.3\%)} \\

\rowcolor{rowVtwoSixty}
~~~~+ A*-Thought-V2-60°
& 31.67\%
& 0:16:57 \textcolor{green}{\scriptsize (-94.6\%)}
& 2:25:56 \textcolor{green}{\scriptsize (-80.3\%)}
& 6:42:07 \textcolor{green}{\scriptsize (-68.7\%)} \\

\rowcolor{rowVtwoNinety}
~~~~+ A*-Thought-V2-90°
& 48.09\%
& 0:16:57 \textcolor{green}{\scriptsize (-94.6\%)}
& 3:47:22 \textcolor{green}{\scriptsize (-69.3\%)}
& 10:18:49 \textcolor{green}{\scriptsize (-51.8\%)} \\

\bottomrule
\end{tabular}%
}
\end{table}

\section{Related Work}

\paragraph{Efficient Reasoning}
Recent work improves reasoning efficiency by shortening explicit CoT trajectories~\cite{recut, s1, cotValve, treeRL, lightthinker, adaR1, yong2025, zhao2025, wu2026, yang2026}. TokenSkip~\cite{tokenskip} prunes low-importance tokens, while A*-Thought~\cite{A*-Thought} combines bidirectional importance scoring with A* search to identify compact reasoning paths. However, such hard pruning may discard useful intermediate information. A*-Thought-V2 instead encodes redundant steps as dense latent representations and interleaves them with retained text, enabling higher-density and more information-preserving compression.

\paragraph{Latent Reasoning}
While traditional CoT improves LLM reasoning, generating lengthy discrete token sequences is computationally expensive. To address this limitation, recent studies have increasingly explored continuous and implicit reasoning paradigms \cite{coconut, CODI, PCCOT, simcot, lcmteam, ConceptLM, LTO, zhu2025, DLCM, CoLaR}. Specifically, SwiReasoning \cite{swireasoning} introduces a dynamic switching mechanism between explicit textual reasoning and latent thinking, achieving a Pareto-superior balance between performance and efficiency. CopT \cite{copt} employs continuous-space verifiers to refine draft answers through on-policy reflection and correction. Following these paradigms, A*-Thought-V2 complements these methods with a trajectory-based criterion for explicit--latent step allocation and step-wise embedding and label forcing, enabling dynamic, variable-length, and information-preserving compression.

\section{Conclusion}

We presented A*-Thought-V2, a dynamics-guided framework that replaces hard CoT pruning with explicit--implicit interleaving. It models reasoning as a PCA-projected hidden-state trajectory, retaining solution-aligned steps as text and compressing deviating steps into latent representations. Directional-angle analysis links six intervals to recurring lexical patterns and characterizes exploration, convergence, and refinement. Step-wise embedding forcing and label forcing supervise latent learning with continuous embeddings and soft vocabulary targets. Across six benchmarks, A*-Thought-V2 improves average accuracy by up to 2.6\% over same-data SFT, relative to the original backbones, the $90^\circ$ variant nearly halves response length and improves ACU by up to $2.29\times$. It also reduces preprocessing time by 94.6\% over A*-Thought and training time by up to 80.3\% over standard SFT. Ablations support the complementary roles of both mechanisms. Representation analyses suggest that latent states form a compact region distinct from textual states, while higher predictive entropy reflects broader soft targets that encourage richer step-level feature learning. Future work will explore reasoning under reinforcement learning.

\bibliography{main}
\bibliographystyle{iclr2027_conference}

\newpage
\appendix

\section{A*-Thought-V2 Algorithm Detail}

\begin{algorithm}
\caption{A*-Thought-V2 algorithm for compressing lengthy CoTs}
\label{alg:a-thought-v2}
\small
\begin{algorithmic}[1]
\State \textbf{Input:}
\State \quad $\mathbf{q}$: question; $\mathbf{t}=\langle\mathbf{t}^{(1)},\ldots,\mathbf{t}^{(N)}\rangle$: segmented CoT; $\mathbf{s}$: solution
\State \quad $\mathcal{H}$: step-level hidden-state extractor; $W_E$: embedding matrix; $\tau\in[0,\pi]$: angle threshold
\State \quad $x_{\mathrm{BOL}},x_{\mathrm{EOL}}$: latent boundary tags; $\epsilon>0$: numerical tolerance

\State \textbf{Output:}
\State \quad $\mathbf{E}_{\mathbf{t}^{\prime}}$: explicit--implicit embeddings; $\mathbf{Y}_{\mathbf{t}^{\prime}}$: aligned hard/soft targets
\State \quad $\mathbf{f}=\langle f_1,\ldots,f_N\rangle$: step-wise compression decisions

\Procedure{A*-Thought-V2}{}
    \Comment{(1) Hidden-state trajectory construction}
    \State $\{\mathbf{h}_{q},\mathbf{h}_{t^{(1)}},\ldots,
    \mathbf{h}_{t^{(N)}},\mathbf{h}_{s}\}
    \gets\mathcal{H}(\mathbf{q},\mathbf{t},\mathbf{s})$
    \State $(P,\boldsymbol{\mu})\gets\operatorname{PCA}_{3}
    (\mathbf{h}_{q},\mathbf{h}_{t^{(1)}},\ldots,
    \mathbf{h}_{t^{(N)}},\mathbf{h}_{s})$
    \For{$x$ in $\{\mathbf{q},\mathbf{t}^{(1)},\ldots,
    \mathbf{t}^{(N)},\mathbf{s}\}$}
        \State $\widetilde{\mathbf{h}}_{x}
        \gets P(\mathbf{h}_{x}-\boldsymbol{\mu})$
    \EndFor
    \State $\mathbf{z}_{0}
    \gets\widetilde{\mathbf{h}}_{s}
    -\widetilde{\mathbf{h}}_{q}$

    \Comment{(2) Dynamics-guided compression decisions}
    \For{$n=1$ to $N$}
        \If{$n=1$}
            \State $\mathbf{z}_{n}
            \gets\widetilde{\mathbf{h}}_{t^{(1)}}
            -\widetilde{\mathbf{h}}_{q}$
        \Else
            \State $\mathbf{z}_{n}
            \gets\widetilde{\mathbf{h}}_{t^{(n)}}
            -\widetilde{\mathbf{h}}_{t^{(n-1)}}$
        \EndIf

        \If{$\lVert\mathbf{z}_{0}\rVert_{2}
        \lVert\mathbf{z}_{n}\rVert_{2}\le\epsilon$}
            \State $f_n\gets1$
            \Comment{retain when the angle is undefined}
        \Else
            \State $\rho_n\gets\operatorname{clip}\!\left(
            \dfrac{\mathbf{z}_{n}^{\top}\mathbf{z}_{0}}
            {\lVert\mathbf{z}_{n}\rVert_{2}
             \lVert\mathbf{z}_{0}\rVert_{2}},-1,1\right)$
            \State $\theta_n\gets\arccos(\rho_n)$;
            $f_n\gets\mathbb{I}[\theta_n\le\tau]$
        \EndIf
    \EndFor

    \Comment{(3) Explicit--implicit sequence construction}
    \State $\mathcal{S}_{\mathbf{t}^{\prime}}
    \gets\langle\rangle$; $b\gets0$

    \For{$n=1$ to $N$}
        \State Write
        $\mathbf{t}^{(n)}
        =\langle x_1^{(n)},\ldots,x_{l_n}^{(n)}\rangle$

        \If{$f_n=0$}
            \If{$b=0$}
                \State $\mathcal{S}_{\mathbf{t}^{\prime}}
                \gets\mathcal{S}_{\mathbf{t}^{\prime}}
                \oplus\operatorname{Tok}(x_{\mathrm{BOL}})$;
                $b\gets1$
            \EndIf

            \State $\mathbf{c}^{(n)}
            \gets\dfrac{1}{l_n}
            \sum_{j=1}^{l_n}W_E[x_j^{(n)},:]$

            \State $\mathbf{y}_{\mathrm{soft}}^{(n)}
            \gets\dfrac{1}{l_n}
            \sum_{j=1}^{l_n}
            \operatorname{onehot}(x_j^{(n)})$

            \State $\mathcal{S}_{\mathbf{t}^{\prime}}
            \gets\mathcal{S}_{\mathbf{t}^{\prime}}
            \oplus
            (\mathbf{c}^{(n)},\mathbf{y}_{\mathrm{soft}}^{(n)})$

        \Else
            \If{$b=1$}
                \State $\mathcal{S}_{\mathbf{t}^{\prime}}
                \gets\mathcal{S}_{\mathbf{t}^{\prime}}
                \oplus\operatorname{Tok}(x_{\mathrm{EOL}})$;
                $b\gets0$
            \EndIf

            \State $\mathcal{S}_{\mathbf{t}^{\prime}}
            \gets\mathcal{S}_{\mathbf{t}^{\prime}}
            \oplus
            \langle
            \operatorname{Tok}(x_1^{(n)}),\ldots,
            \operatorname{Tok}(x_{l_n}^{(n)})
            \rangle$
        \EndIf
    \EndFor

    \If{$b=1$}
        \State $\mathcal{S}_{\mathbf{t}^{\prime}}
        \gets\mathcal{S}_{\mathbf{t}^{\prime}}
        \oplus\operatorname{Tok}(x_{\mathrm{EOL}})$
    \EndIf

    \State Split $\mathcal{S}_{\mathbf{t}^{\prime}}$
    into $\mathbf{E}_{\mathbf{t}^{\prime}}$
    and $\mathbf{Y}_{\mathbf{t}^{\prime}}$

    \State \Return
    $\mathbf{E}_{\mathbf{t}^{\prime}},
    \mathbf{Y}_{\mathbf{t}^{\prime}},\mathbf{f}$
\EndProcedure
\end{algorithmic}
\end{algorithm}

\section{Geometric and Directional Dynamics Analysis of the LLM Reasoning Representation Trajectory}
\label{app:dynamics-analysis}

\subsection{Angle Threshold and Reasoning Semantic Analysis}
\label{app:angle-semantic-analysis}

\begin{table*}[ht]
\centering
\small
\setlength{\tabcolsep}{5pt}
\renewcommand{\arraystretch}{1.2}
\caption{Semantic keyword groups and proportions associated with directional-angle intervals.}
\label{tab:angle_semantic_keywords}
\begin{tabularx}{\textwidth}{
    >{\centering\arraybackslash}p{1.7cm}
    >{\centering\arraybackslash}p{1.2cm}
    >{\raggedright\arraybackslash}p{3.1cm}
    >{\raggedright\arraybackslash}X
}
\toprule
\textbf{Angle Range} &
\textbf{Share} &
\textbf{Interval Semantics} &
\textbf{Keywords} \\
\midrule
\rowcolor{anglebin1}
$[~0^\circ,~30^\circ)$ &
$7.57\%$ &
Direct reduction / Conclusion &
\texttt{expand}, \texttt{substitute}, \texttt{hence},
\texttt{explain}, \texttt{simplify}, \texttt{overlooked},
\texttt{remaining}, \texttt{equivalently}, \texttt{we are asked},
\texttt{hours}, \texttt{answer boxed frac}, \texttt{is sqrt},
\texttt{wait compute}, \texttt{days}, \texttt{total so},
\texttt{initially}, \texttt{sum sum} \\
\addlinespace[3pt]
\rowcolor{anglebin2}
$[~30^\circ,~60^\circ)$ &
$18.46\%$ &
Routine derivation / Checking &
\texttt{thus}, \texttt{case}, \texttt{yields}, \texttt{satisfies},
\texttt{equals}, \texttt{alternatively}, \texttt{check},
\texttt{boxed frac sqrt}, \texttt{answer frac}, \texttt{bp},
\texttt{for product}, \texttt{compute cos}, \texttt{alpha},
\texttt{invariant}, \texttt{to mod}, \texttt{sin cos sin},
\texttt{cos sin cos} \\
\addlinespace[3pt]
\rowcolor{anglebin3}
$[~60^\circ,~90^\circ)$ &
$24.72\%$ &
Mixed execution / Transition &
\texttt{boxed}, \texttt{compute}, \texttt{solve}, \texttt{then},
\texttt{actually}, \texttt{evaluate}, \texttt{indeed},
\texttt{window}, \texttt{thus done}, \texttt{thus solution is},
\texttt{answer is correct}, \texttt{and boxed},
\texttt{decomposition}, \texttt{thus our}, \texttt{then that's},
\texttt{thus answer seems}, \texttt{answer stands} \\
\addlinespace[3pt]
\rowcolor{anglebin4}
$[~90^\circ,~120^\circ)$ &
$24.50\%$ &
Condition checking / Revision &
\texttt{consistent}, \texttt{verify}, \texttt{holds}, \texttt{wait},
\texttt{wrong}, \texttt{unless}, \texttt{exclude}, \texttt{implies},
\texttt{given the problem}, \texttt{no further}, \texttt{i'll also},
\texttt{accept}, \texttt{answer yes}, \texttt{they might want},
\texttt{but i'll}, \texttt{implicitly}, \texttt{the blank},
\texttt{answer so} \\
\addlinespace[3pt]
\rowcolor{anglebin5}
$[~120^\circ,~150^\circ)$ &
$17.80\%$ &
Reinterpretation / Correction &
\texttt{present}, \texttt{interpret}, \texttt{mistake}, \texttt{but},
\texttt{approximately}, \texttt{eliminate}, \texttt{rigorous},
\texttt{however}, \texttt{the interpretation},
\texttt{the condition holds}, \texttt{any additional},
\texttt{let's read the}, \texttt{at two points},
\texttt{an alternative}, \texttt{but they said},
\texttt{exactly what we}, \texttt{can simplify},
\texttt{but perhaps the} \\
\addlinespace[3pt]
\rowcolor{anglebin6}
$[~150^\circ,~180^\circ]$ &
$6.96\%$ &
Reconsideration / Branching &
\texttt{consider}, \texttt{maybe}, \texttt{confirm}, \texttt{ensure},
\texttt{perhaps}, \texttt{reconsider}, \texttt{tangent to the},
\texttt{on side}, \texttt{is inside the}, \texttt{the lateral},
\texttt{is tangent to}, \texttt{touches}, \texttt{wait is there},
\texttt{sphere}, \texttt{wrote}, \texttt{intersecting} \\
\bottomrule
\end{tabularx}
\end{table*}

\begin{figure}[ht]
    \centering
    \includegraphics[width=0.9\linewidth]
    {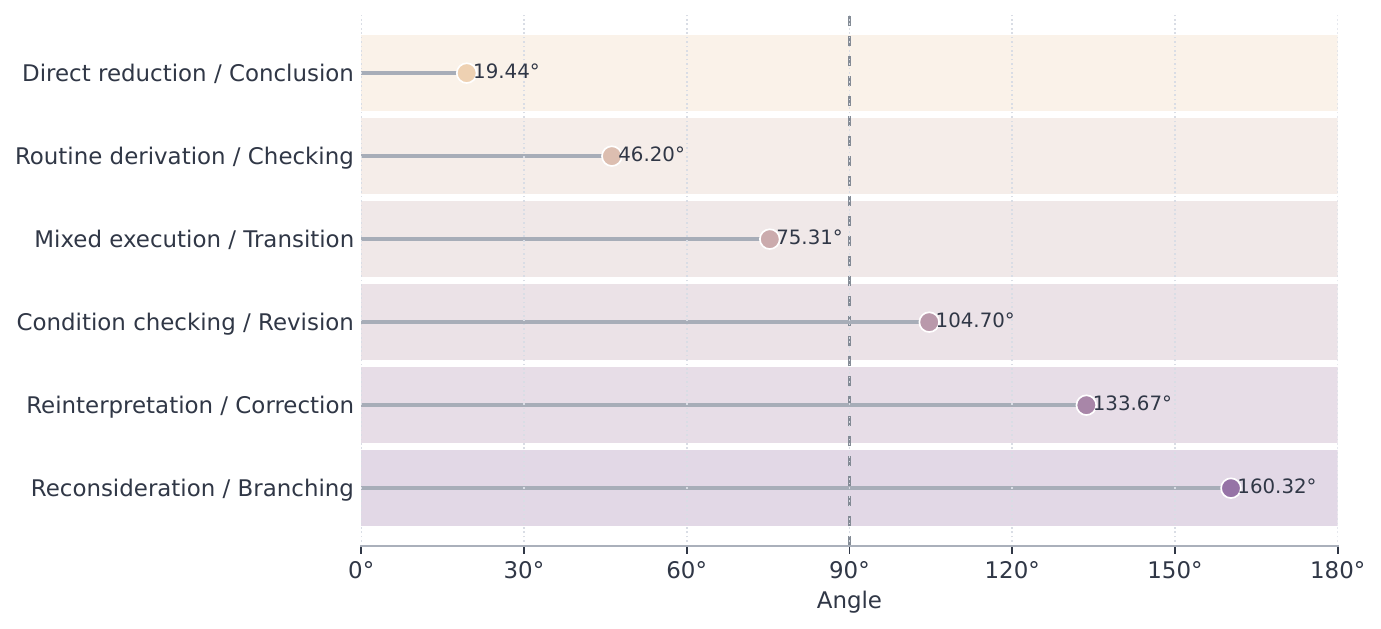}
    % \caption{Mean directional angles of thought steps within six directional-angle intervals}
    \caption{Visual summary of the six predefined $30^\circ$ directional-angle intervals.}
    \label{fig:semantic_group_angles}
\end{figure}

To characterize how lexical patterns vary with trajectory direction, we partition thought steps into six predefined $30^\circ$ directional-angle intervals and summarize the recurring lexical markers observed within each interval (Table~\ref{tab:angle_semantic_keywords}). Lower-angle intervals contain execution and reduction markers, such as \textit{hence}, \textit{simplify}, \textit{thus}, and \textit{compute}, whereas higher-angle intervals contain checking, revision, and branch-exploration markers, such as \textit{wait}, \textit{however}, \textit{maybe}, and \textit{wait is there}. Figure~\ref{fig:semantic_group_angles} provides a visual reference for the six bins by displaying their within-bin mean angles. Together, the lexical patterns in Table~\ref{tab:angle_semantic_keywords} provide a descriptive view of how reasoning expressions vary across directional-angle intervals.

\newpage
\subsection{Case Study of Step-wise Directional Dynamics}
\label{app:step-dynamics-case}

\begin{figure}[ht]
	\centering
	\includegraphics[width=0.9\linewidth]{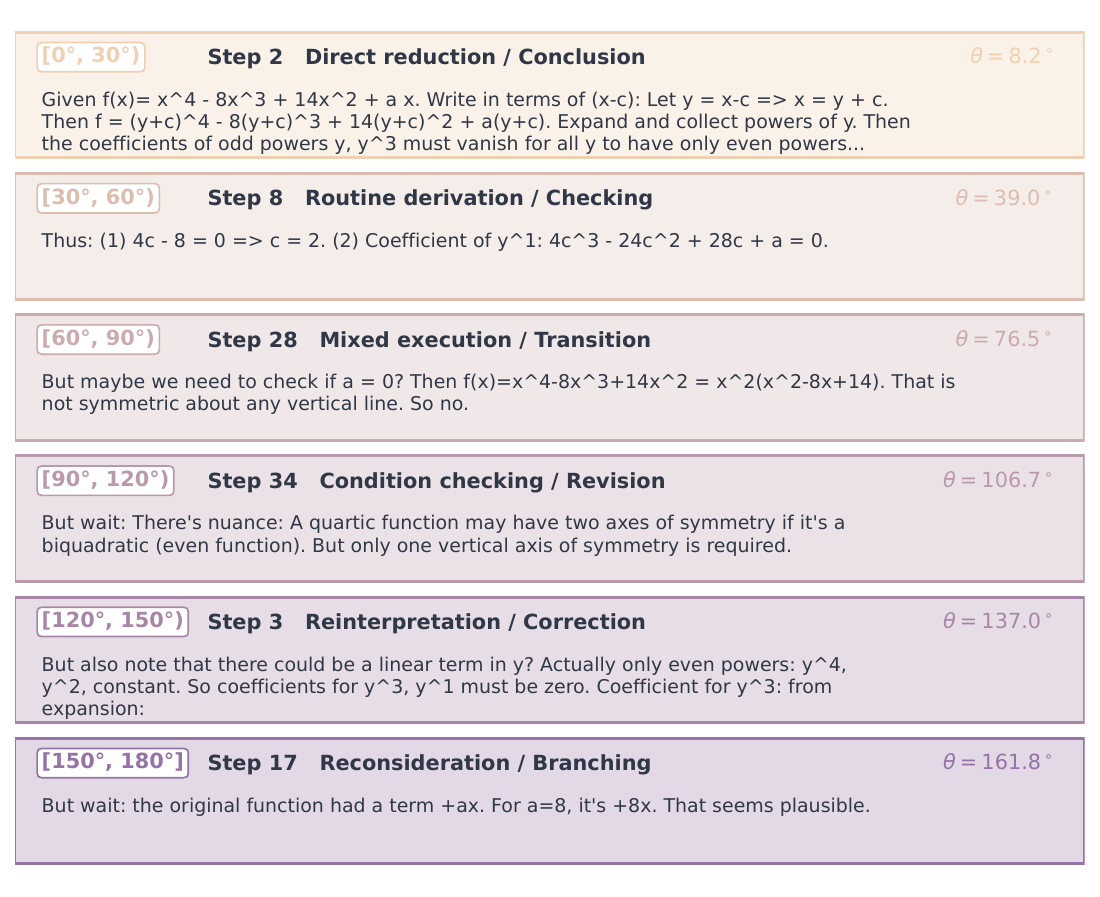}
	\caption{Original reasoning content of the six steps marked in Figure~\ref{fig:pca}(c), covering consecutive directional-angle intervals within the same trajectory.}
    \label{fig:case_dynamics}
\end{figure}

Figure~\ref{fig:case_dynamics} expands the steps marked in Figure~\ref{fig:pca}(c) into their original reasoning content. These six steps come from the same trajectory and cover consecutive $30^\circ$ intervals. Smaller-angle steps primarily perform direct algebraic reduction, whereas larger-angle steps more frequently involve condition checking, reinterpretation, and alternative branches. The examples are ordered by angle interval rather than generation order, while their step indices indicate their original positions in the trajectory.

\newpage
\section{Analysis}
\label{app:analysis}

\subsection{Effect of CoT-PCA Extractor Size}

\begin{figure}[ht]
    \centering
    \subfigcapskip=-3pt
    \subfigure[Qwen3.5-0.8B]{
        \includegraphics[width=0.6\linewidth]{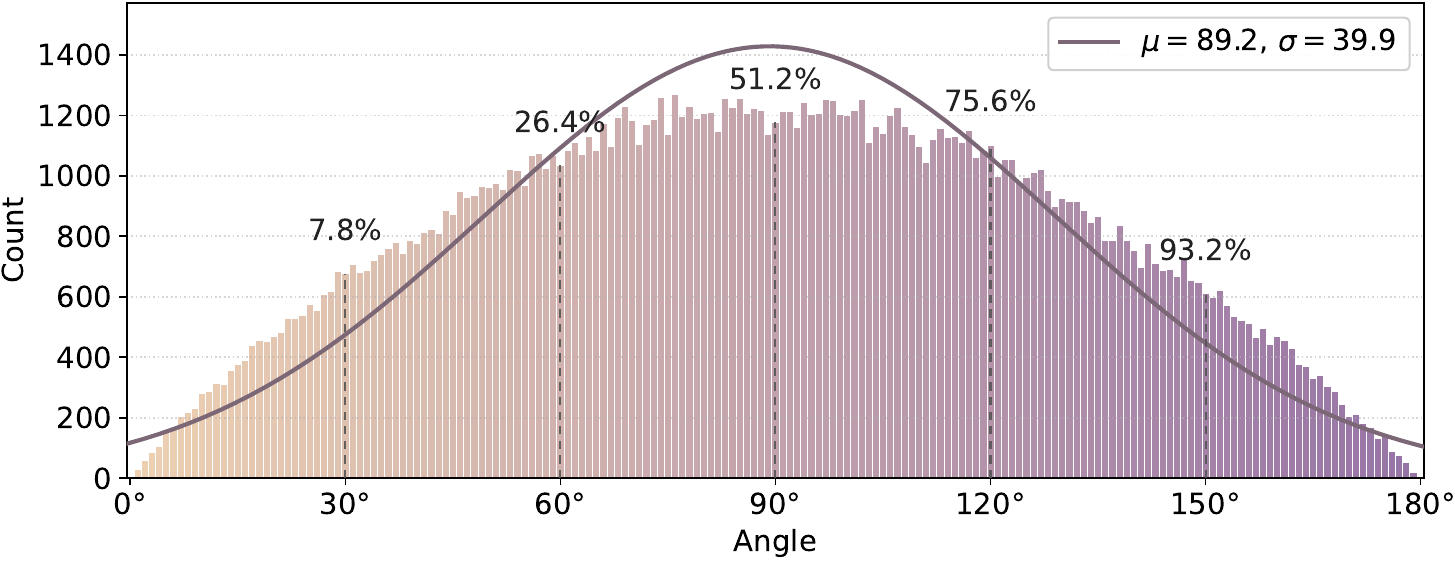}}
    \subfigure[Qwen3.5-2B]{
        \includegraphics[width=0.6\linewidth]{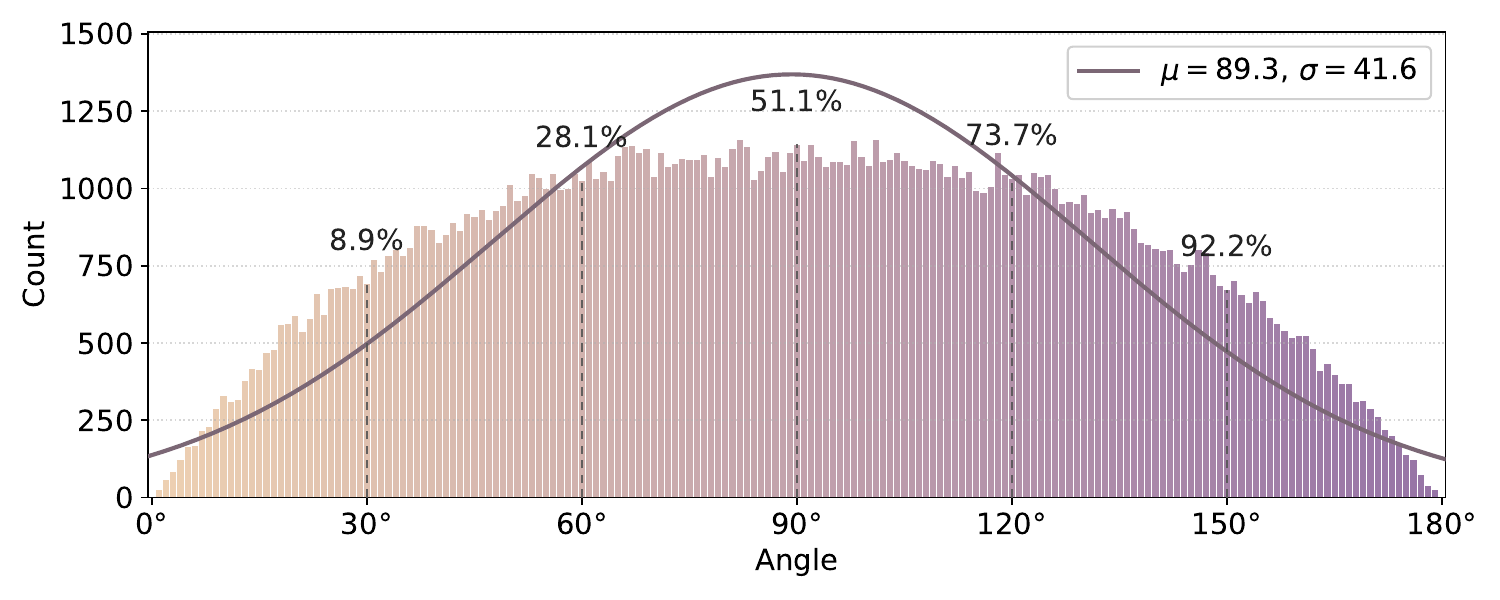}}
    \subfigure[Qwen3.5-4B]{
        \includegraphics[width=0.6\linewidth]{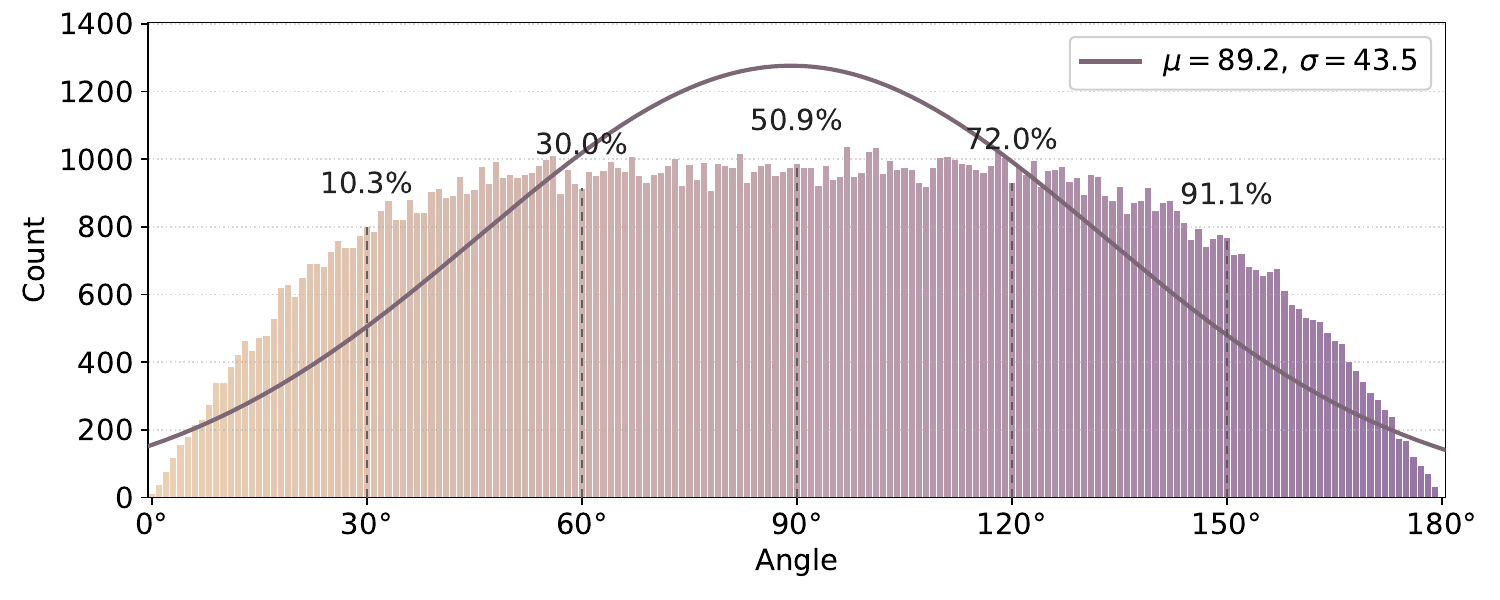}}
    \subfigure[Qwen3.5-9B]{
        \includegraphics[width=0.6\linewidth]{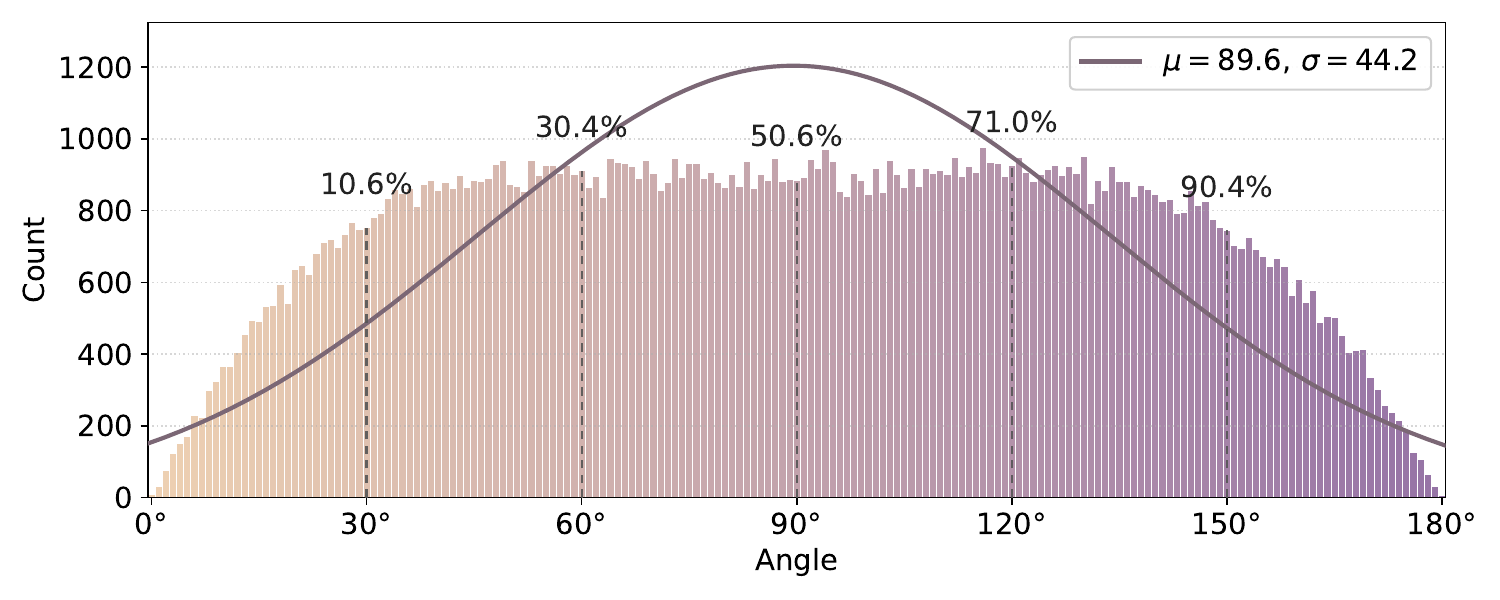}}
    \caption{Directional-angle distributions produced by CoT-PCA extractors of different sizes.}
    \label{fig:ablation_pca_angle}
\end{figure}

We investigate how the representation model used to extract CoT hidden states affects CoT-PCA. Specifically, we compare four Qwen3.5 extractors with 0.8B, 2B, 4B, and 9B parameters. As shown in Figure~\ref{fig:ablation_pca_angle}, different extractor sizes produce slightly different angle distributions and therefore affect the interpretation of the angle threshold. However, downstream performance shows no clear gap across model sizes, indicating that CoT-PCA is not highly sensitive to extractor scale. Considering both effectiveness and computational efficiency, we use Qwen3.5-0.8B as the default extractor.

% \subsection{Effect of Maximum Latent Count}
% \label{app:analysis_max_latent_count}

% We further investigate the maximum latent count, which limits the number of latent segments that may occur within a reasoning trajectory. Figure~\ref{fig:ablation_latent_count} reports its effects on accuracy and response length across AIME2024 and GPQA-Diamond. This experiment isolates the influence of text--latent switching frequency from the length of each latent segment. The resulting latent-count distribution further shows how frequently the model uses latent reasoning under different count constraints.

% \begin{figure}[ht]
%     \centering
%     \subfigcapskip=-3pt
%     \subfigure[Accuracy]{
%         \includegraphics[width=0.4\linewidth]{figure/max_latent_count_acc.pdf}}%
%     \subfigure[Length]{
%         \includegraphics[width=0.4\linewidth]{figure/max_latent_count_tokens.pdf}}%
%     \par\smallskip
%     \subfigure[Latent Count Distribution]{
%         \includegraphics[width=0.8\linewidth]{figure/latent_count_distribution.pdf}}
%     \caption{Effect of maximum latent count on accuracy, response length, and latent-count distribution.}
%     \label{fig:ablation_latent_count}
% \end{figure}

\newpage
\section{Hyperparameters}
\label{app:hyperparameters}

We report the training and inference configurations in Tables~\ref{tab:training_hyperparams} and~\ref{tab:inference_hyperparams}, respectively. All backbone models use the same optimization settings, while method-specific hyperparameters are configured separately.

\begin{table}[ht]
\centering
\caption{Training configuration.}
\label{tab:training_hyperparams}
\setlength{\tabcolsep}{8pt}
\resizebox{\textwidth}{!}{%
\begin{tabular}{l cccc}
\toprule
\textbf{Hyperparameter} & \multicolumn{4}{c}{\textbf{Value}} \\
\midrule

\multicolumn{5}{l}{\textbf{Foundation}} \\
\quad Base model
& \multicolumn{2}{c}{Qwen3.5-9B}
& \multicolumn{2}{c}{Qwen3.6-27B} \\
\quad GPUs
& \multicolumn{4}{c}{8 $\times$ NVIDIA A100 (80GB)} \\
\quad Device batch size
& \multicolumn{4}{c}{1} \\
\quad Gradient accumulation steps
& \multicolumn{4}{c}{8} \\
\quad Total batch size
& \multicolumn{4}{c}{64} \\
\quad Cutoff length
& \multicolumn{4}{c}{20480} \\
\quad Optimizer
& \multicolumn{4}{c}{AdamW} \\
\quad Learning rate
& \multicolumn{4}{c}{$1.0 \times 10^{-5}$} \\
\quad Warmup ratio
& \multicolumn{4}{c}{0.1} \\
\quad Epochs
& \multicolumn{4}{c}{3} \\

\midrule
\multicolumn{5}{l}{\textbf{A*-Thought}} \\
\quad $\alpha$
& \multicolumn{4}{c}{0.5} \\
\quad $\beta$
& \multicolumn{4}{c}{0.1} \\
\quad $k_{\text{min}}$
& \multicolumn{4}{c}{25} \\
\quad $k_{\text{max}}$
& \multicolumn{4}{c}{40} \\

\midrule
\multicolumn{5}{l}{\textbf{A*-Thought-V2 (Ours)}} \\
\quad Angle threshold
& $60^\circ$ & $90^\circ$
& $60^\circ$ & $90^\circ$ \\
\quad Latent weight $\lambda$
& 0.2 & 0.3
& 0.1 & 0.1 \\
\quad Number of added latent tokens
& \multicolumn{4}{c}{256} \\

\bottomrule
\end{tabular}%
}
\end{table}

\begin{table}[htbp]
\centering
\caption{Inference configuration.}
\label{tab:inference_hyperparams}
\setlength{\tabcolsep}{24pt}
\begin{tabular}{lc}
\toprule
\textbf{Hyperparameter} & \textbf{Value} \\
\midrule

\multicolumn{2}{l}{\textbf{Foundation}} \\
\quad Temperature & 1.0 \\
\quad Top-$p$ & 0.95 \\
\quad Max tokens & 81920 \\
\quad Repeat & 4 \\

\midrule
\multicolumn{2}{l}{\textbf{A*-Thought-V2 (Ours)}} \\
\quad Max latent count & 4 \\
\quad Max latent length & 32 \\

\bottomrule
\end{tabular}
\end{table}

The latent weight $\lambda$ controls the contribution of the latent-token objective. We tune $\lambda$ through a grid search with a step size of $0.1$ and monitor latent-token generation across benchmarks. A small $\lambda$ may fail to activate latent generation, whereas a large value may produce excessive latent tokens and lead to overfitting. We therefore select the smallest value that reliably induces latent-token generation without excessive outputs.

\section{Case Study of the A*-Thought-V2 Trajectories}
\label{app:case}

\begin{figure}[H]
  \vspace*{-8mm}
  \centering
  \begin{minipage}[t]{0.98\textwidth}
    \begin{tcolorbox}[
      enhanced,
      colback=casequestionbg,
      colframe=caseborder,
      boxrule=0.7pt,
      arc=1.6mm,
      left=2.7mm,
      right=2.7mm,
      top=2.0mm,
      bottom=2.0mm,
      title={\textbf{Question}},
      coltitle=black,
      fonttitle=\small,
      attach boxed title to top left={xshift=2mm,yshift*=-1.1mm},
      boxed title style={
        colback=casequestiontitle,
        colframe=caseborder,
        boxrule=0.7pt,
        arc=1.2mm
      }
    ]
    \small
    Find the greatest integer $n$ such that the cubic polynomial
    \[
      x^{3}-\frac{n}{6}x^{2}+(n-11)x-400
    \]
    has roots $\alpha^{2}$, $\beta^{2}$, and $\gamma^{2}$, where $\alpha$,
    $\beta$, and $\gamma$ are complex numbers, and there are exactly seven
    different possible values for $\alpha+\beta+\gamma$.
    \end{tcolorbox}

    \vspace{2mm}

    \begin{tcolorbox}[
      enhanced,
      colback=white,
      colframe=caseborder,
      boxrule=0.7pt,
      arc=1.6mm,
      left=2.7mm,
      right=2.7mm,
      top=2.0mm,
      bottom=2.0mm,
      title={\textbf{Output}},
      coltitle=black,
      fonttitle=\small,
      attach boxed title to top left={xshift=2mm,yshift*=-1.1mm},
      boxed title style={
        colback=caseoutputtitle,
        colframe=caseborder,
        boxrule=0.7pt,
        arc=1.2mm
      }
    ]
    \footnotesize
    \setlength{\parindent}{0pt}
    \setlength{\parskip}{2.2pt}

    \caseRawTag{<think>}

    Let $r_1,r_2,r_3$ be the roots, so $\alpha^2=r_1$, $\beta^2=r_2$,
    $\gamma^2=r_3$. By Vieta's formulas,
    \[
      r_1+r_2+r_3=\frac{n}{6},
      \qquad
      r_1r_2+r_1r_3+r_2r_3=n-11,
      \qquad
      r_1r_2r_3=400 .
    \]
    Since the product is nonzero, no root vanishes. Fixing square roots
    $a,b,c$ of $r_1,r_2,r_3$, every admissible sum is
    $\alpha+\beta+\gamma=\pm a\pm b\pm c$, giving at most eight values.
    \caseExplicitEllipsis

    The eight sign patterns form four pairs of opposites, so the set of sums
    is symmetric under negation. If two non-opposite patterns agreed, their
    opposites would agree too and at most six values would remain. Hence
    exactly seven values occur precisely when some sum vanishes,
    $\pm a\pm b\pm c=0$, with all three coefficients nonzero. Squaring this
    relation gives the symmetric condition
    \[
      (r_1+r_2+r_3)^{2}=4\left(r_1r_2+r_1r_3+r_2r_3\right).
    \]
    \caseExplicitEllipsis

    A relation with a vanishing coefficient does not produce seven values.
    If $a=-b$, the sums collapse to
    $\{\,c,\,-c,\,2a\pm c,\,-2a\pm c\,\}$, that is, only six values, and
    $a=b$ leaves even fewer. So the vanishing combination must involve all
    three square roots, and after adjusting signs we may assume $a=b+c$.
    \caseExplicitEllipsis

    Substituting the Vieta expressions yields
    \[
      \left(\frac{n}{6}\right)^{2}=4(n-11)
      \;\Longrightarrow\;
      n^{2}-144n+1584=0
      \;\Longrightarrow\;
      n=\frac{144\pm 120}{2},
    \]
    so $n=132$ or $n=12$, and the greatest is $132$.

    Thus the answer should be $132$.

    \caseLatentSpan{<latent><latent_1><latent_2></latent>}

    as we did. So we have exactly seven different possible values for
    $\alpha+\beta+\gamma$. So $n=132$ works.

    \caseLatentSpan{<latent><latent_3><latent_4><latent_5><latent_6></latent>}

    This is exactly the condition.

    \caseLatentSpan{<latent><latent_7><latent_8></latent>}

    For $n=132$ the polynomial factors as
    $x^{3}-22x^{2}+121x-400=(x-16)(x^{2}-6x+25)$, with roots $16$ and
    $3\pm4i$. Choosing $a=4$, $b=-2-i$, $c=-2+i$ gives $a+b+c=0$, and the
    eight signed sums are $8$, $4\pm2i$, $0$, $-4\pm2i$, $-8$ --- exactly
    seven distinct values.

    \caseRawTag{</think>}

    \vspace{0.8mm}
    \begin{tcolorbox}[
      enhanced,
      colback=casegreenbg,
      colframe=casegreen,
      boxrule=0.7pt,
      arc=1.2mm,
      left=2mm,
      right=2mm,
      top=1.4mm,
      bottom=1.4mm
    ]
      \textbf{Final answer.}
      The seven-sum condition is equivalent to $n^{2}-144n+1584=0$, whose
      solutions are $n=132$ and $n=12$, so the greatest such integer is
      \[
        \boxed{132}.
      \]
    \end{tcolorbox}
    \end{tcolorbox}

    \caption{Case study of A*-Thought-V2-Qwen3.5-9B on AIME2026-ID25.
    Representative explicit reasoning is retained before ellipses, while all
    latent markers in the original output are retained and
    \caseLatentSpan{highlighted in purple}.}
    \label{fig:a-thought-v2-qwen35-9b-aime2026-id25}
  \end{minipage}
\end{figure}

\begin{figure}[H]
  \centering
  \begin{minipage}[t]{0.98\textwidth}
    \begin{tcolorbox}[
      enhanced,
      colback=casequestionbg,
      colframe=caseborder,
      boxrule=0.7pt,
      arc=1.6mm,
      left=2.7mm,
      right=2.7mm,
      top=2.0mm,
      bottom=2.0mm,
      title={\textbf{Question}},
      coltitle=black,
      fonttitle=\small,
      attach boxed title to top left={xshift=2mm,yshift*=-1.1mm},
      boxed title style={
        colback=casequestiontitle,
        colframe=caseborder,
        boxrule=0.7pt,
        arc=1.2mm
      }
    ]
    \small
    Alice and Bob play the following game. A stack of $n$ tokens lies before
    them. The players take turns with Alice going first. On each turn, the
    player removes either $1$ token or $4$ tokens from the stack. Whoever
    removes the last token wins. Find the number of positive integers $n$
    less than or equal to $2024$ for which there exists a strategy for Bob
    that guarantees that Bob will win the game regardless of Alice's play.
    \end{tcolorbox}

    \vspace{2mm}

    \begin{tcolorbox}[
      enhanced,
      colback=white,
      colframe=caseborder,
      boxrule=0.7pt,
      arc=1.6mm,
      left=2.7mm,
      right=2.7mm,
      top=2.0mm,
      bottom=2.0mm,
      title={\textbf{Output}},
      coltitle=black,
      fonttitle=\small,
      attach boxed title to top left={xshift=2mm,yshift*=-1.1mm},
      boxed title style={
        colback=caseoutputtitle,
        colframe=caseborder,
        boxrule=0.7pt,
        arc=1.2mm
      }
    ]
    \footnotesize
    \setlength{\parindent}{0pt}
    \setlength{\parskip}{2.2pt}

    \caseRawTag{<think>}

    This is an impartial game under normal play: from a stack of $n$ tokens a
    player may move to $n-1$ or $n-4$, and the player taking the last token
    wins. Since Alice moves first, Bob has a winning strategy exactly when
    the starting position is losing for the player about to move, i.e.\ a
    $P$-position. So the task is to count $P$-positions with $n\le 2024$.

    \caseLatentSpan{<latent><latent_1><latent_2><latent_3><latent_4><latent_5></latent>}

    The Grundy values satisfy $g(0)=0$ and
    \[
      g(n)=\operatorname{mex}\{\,g(n-1),\,g(n-4)\,\},
    \]
    where the term $g(n-4)$ is present only for $n\ge4$. Computing the first
    values gives
    \[
      \begin{array}{c|cccccccccccccccc}
        n    & 0&1&2&3&4&5&6&7&8&9&10&11&12&13&14&15\\\hline
        g(n) & 0&1&0&1&2&0&1&0&1&2&0 &1 &0 &1 &2 &0
      \end{array}
    \]
    The pattern is periodic with period $5$: $g(n)=0$ when $n\equiv0,2$,
    $g(n)=1$ when $n\equiv1,3$, and $g(n)=2$ when $n\equiv4 \pmod 5$. Hence
    the $P$-positions appear to be exactly $n\equiv0$ or $2 \pmod 5$.
    \caseExplicitEllipsis

    \caseLatentSpan{<latent><latent_6></latent>}

    The pattern can be confirmed by induction rather than trusted from a
    finite table. Classify $P=\{n\equiv0,2\}$ and $N=\{n\equiv1,3,4\}$
    modulo $5$. From $n\equiv0$ the two moves land on $n-1\equiv4$ and
    $n-4\equiv1$, both in $N$; from $n\equiv2$ they land on $n-1\equiv1$ and
    $n-4\equiv3$, again both in $N$. Conversely, every $N$-position has a
    move into $P$: from $n\equiv1$ or $n\equiv3$ remove one token to reach
    $n\equiv0$ or $n\equiv2$, and from $n\equiv4$ remove four tokens to reach
    $n\equiv0$. These two properties are exactly the characterisation of
    $P$- and $N$-positions, so the classification is correct for all $n$.
    \caseExplicitEllipsis

    Small cases agree with it: $n=1,3,4$ are wins for Alice (for $n=4$ she
    simply takes all four tokens), while $n=2$ and $n=5$ are wins for Bob.
    For $n=5$ every Alice move leaves $4$ or $1$ tokens, and in both cases
    Bob finishes the stack.

    \caseLatentSpan{<latent><latent_7><latent_8><latent_9></latent>}

    It remains to count the integers $1\le n\le 2024$ with $n\equiv0$ or
    $2\pmod 5$. Since $2024=5\cdot404+4$, the multiples of $5$ are
    $5,10,\dots,2020$, giving $2020/5=404$ values, and the residues
    $2 \pmod 5$ are $2,7,\dots,2022$, giving $(2022-2)/5+1=405$ values. The
    two families are disjoint, so the total is
    \[
      404+405=809 .
    \]
    \caseExplicitEllipsis

    \caseRawTag{</think>}

    \vspace{0.8mm}
    \begin{tcolorbox}[
      enhanced,
      colback=casegreenbg,
      colframe=casegreen,
      boxrule=0.7pt,
      arc=1.2mm,
      left=2mm,
      right=2mm,
      top=1.4mm,
      bottom=1.4mm
    ]
      \textbf{Final answer.}
      Bob wins precisely when $n\equiv0$ or $2\pmod 5$. Among $n\le 2024$
      there are $404$ multiples of $5$ and $405$ integers congruent to $2$,
      so the count is
      \[
        404+405=\boxed{809}.
      \]
    \end{tcolorbox}
    \end{tcolorbox}

    \caption{Case study of A*-Thought-V2-Qwen3.6-27B on AIME2024-ID68.
    Representative explicit reasoning is retained before ellipses, while all
    latent markers in the original output are retained and \caseLatentSpan{highlighted in purple}.}
    \label{fig:a-thought-v2-case-27b-aime2024-id68}
  \end{minipage}
\end{figure}

\section{Ethics Statement}

This work studies the computational efficiency of language-model reasoning and does not introduce a new user-facing deployment. The method inherits the factuality, bias, privacy, and misuse risks of its underlying models and training data.

\section{Reproducibility Statement}

The paper specifies the trajectory criterion, compression threshold, embedding and label objectives, inference mode switch, evaluation metrics, hardware, and method-specific hyperparameters. Tables~\ref{tab:training_hyperparams} and~\ref{tab:inference_hyperparams} report the configurations used in the experiments, while the main results and ablations provide the corresponding accuracy, length, ACU, preprocessing-time, and training-time measurements.

\section{The Use of Large Language Models}

A large language model was used to assist with language polishing. It did not generate or modify the experimental measurements, literature citations, or figure and table contents.

\end{document}